\documentclass[final,3p,times,twocolumn]{elsarticle}

\usepackage{amssymb}
\usepackage{amsmath}

\usepackage{algorithmic}
\usepackage{graphicx}
\usepackage{textcomp}
\usepackage{xcolor}
\usepackage{algorithm}
\usepackage{algorithmic}
\usepackage{amsmath}

\usepackage{amsfonts,amssymb} 
\usepackage{bm}
\usepackage{multirow}
\usepackage{enumitem}
\usepackage{amsmath}
\usepackage{svg}
\usepackage{subfig}
\usepackage{booktabs}
\DeclareMathOperator*{\argmax}{arg\,max}
\DeclareMathOperator*{\argmin}{arg\,min}
\usepackage{hyperref}
\usepackage{wrapfig}

\usepackage{soul}
\sethlcolor{yellow}
\soulregister\cite7       %
\soulregister\ref7        %

\journal{Pattern Recognition}

\begin{document}

\begin{frontmatter}

\title{Label Ensemble Perspective for Reduced-Complexity Graph Node Classification under Label Noise}

\author[1,2]{Rui Zhao}
\author[1,2]{Bin Shi\corref{cor1}}
\author[1,3]{Zhiming Liang}
\author[1,3]{Jianfei Ruan}
\author[1,3]{Kai Sun}
\author[3,4]{Bo Dong}
\author[5]{Lu Lin}

\cortext[cor1]{Corresponding author. 
Email: shibin@xjtu.edu.cn}

\affiliation[1]{organization={School of Computer Science and Technology},%
            addressline={Xi’an Jiaotong University}, 
            city={Xi'an},
            postcode={710049}, 
            country={China}}
\affiliation[2]{organization={National Engineering Lab for Big Data Analytics},%
            addressline={Xi’an Jiaotong University}, 
            city={Xi'an},
            postcode={710049}, 
            country={China}}
\affiliation[3]{organization={Key Laboratory of Intelligent Networks and Network Security},%
            addressline={Ministry of Education}, 
            city={Xi'an},
            postcode={710049}, 
            country={China}}
\affiliation[4]{organization={School of Distance Education},%
            addressline={Xi’an Jiaotong University}, 
            city={Xi'an},
            postcode={710049}, 
            country={China}}
\affiliation[5]{organization={The Pennsylvania State University},
            city={University Park},
            postcode={PA 16802},
            country={USA}
}
\begin{abstract}
Graph Neural Networks (GNNs) have been widely employed for semi-supervised node classification tasks on graphs. However, the performance of GNNs is significantly affected by \emph{label noise}, that is, a small amount of incorrectly labeled nodes can substantially misguide model training. Mainstream solutions define node classification with label noise (NCLN) as a \emph{reliable labeling} task, often introducing node similarity with quadratic computational complexity to more accurately assess label reliability. To this end, in this paper, we introduce the Label Ensemble Graph Neural Network (LEGNN), a lower complexity method for robust GNNs training against label noise. LEGNN reframes NCLN as a label ensemble task, gathering informative multiple labels instead of constructing a single reliable label, avoiding high-complexity computations for reliability assessment. Specifically, LEGNN conducts a two-step process: bootstrapping neighboring contexts and robust learning with gathered multiple labels. In the former step, we apply random neighbor masks for each node and gather the predicted labels as a high-probability label set. This mitigates the impact of inaccurately labeled neighbors and diversifies the label set. In the latter step, we utilize a partial label learning based strategy to aggregate the high-probability label information for model training. Additionally, we symmetrically gather a low-probability label set to counteract potential noise from the high-probability label set. Extensive experiments on six datasets demonstrate that LEGNN achieves strong classification performance while maintaining computational efficiency. Moreover, it exhibits good scalability on datasets with over one hundred thousand nodes and one million edges.
\end{abstract}


\end{frontmatter}

\section{Introduction}
Graph, as a ubiquitous data form in real world, can represent a variety of relational structures, such as social networks \cite{leng2022incorporating} and transportation systems \cite{lei2022modeling}. Graph Neural Networks (GNNs) have emerged as powerful tools for capturing and analyzing graph-structured data, gaining significant attention in recent years \cite{zhang2020robust}. GNNs typically employ a message-passing mechanism that enables nodes to iteratively update their representations by aggregating information from neighboring nodes \cite{kipf2017semi}. This mechanism enables GNNs to excel in \emph{semi-supervised} node classification tasks \cite{zhang2020robust} by allowing a few labeled nodes to propagate their knowledge to numerous unlabeled nodes.

\begin{figure}[t]
    \centering
    \includegraphics[width=0.98\columnwidth]{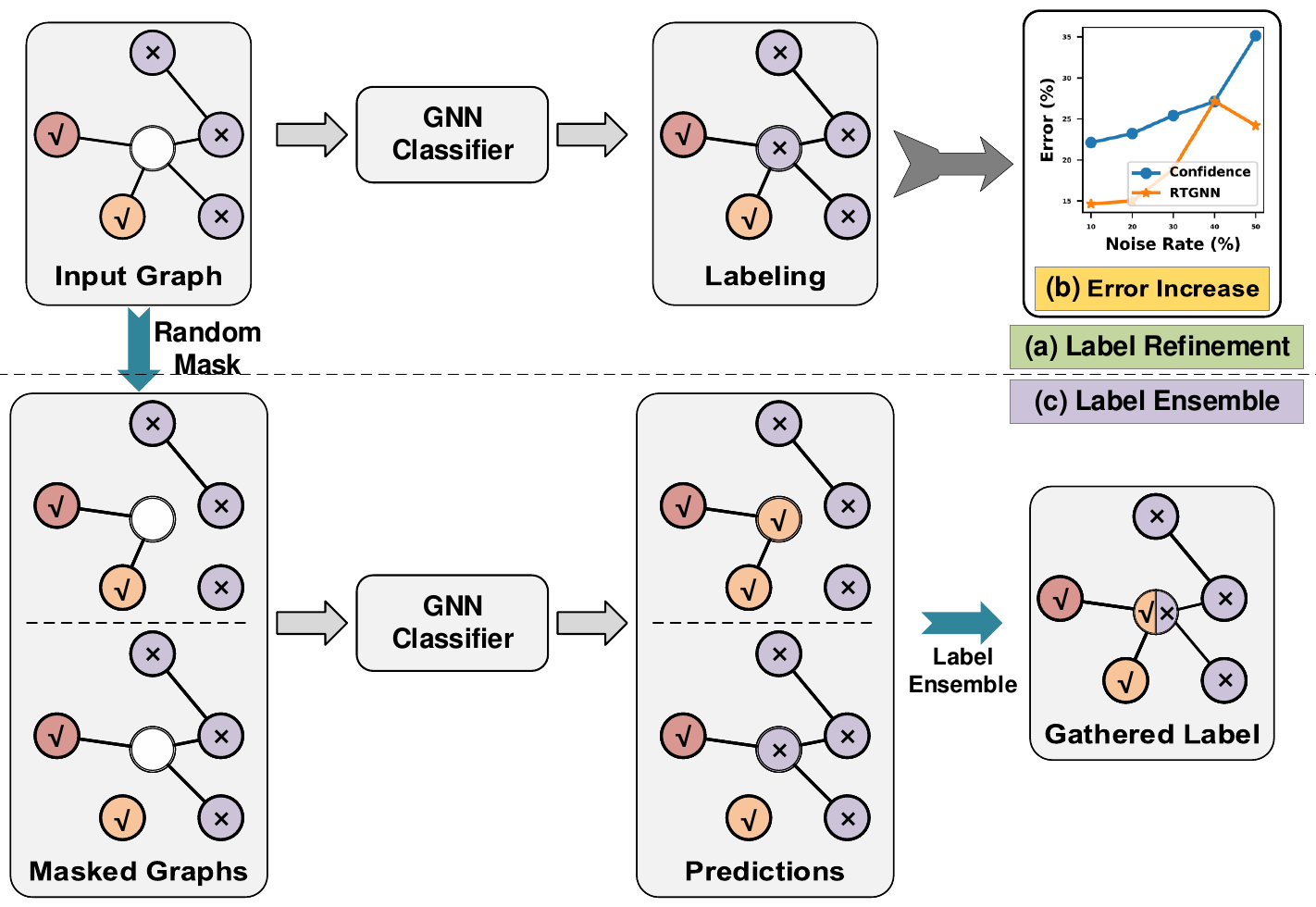}
    \caption{Comparison of our label ensemble method with existing reliable labeling methods. Node colors represent labels; symbols ``\checkmark'' and ``×'' indicate correct and incorrect labels, respectively. Intuitively, our method avoids directly mistaking erroneous labels as correct by label ensemble. In (b), RTGNN \cite{qian2023robust} and Confidence are reliable labeling strategies, with experimental details consistent with those in Section \ref{sec:em}.}
    \label{fig:frameintro}
\end{figure}

Conventional semi-supervised node classification methods usually assume that labeled data is sufficiently clean, meaning that the labeled nodes are correctly annotated \cite{liu2025noisy}. However, such an assumption may not always hold in the real world. For instance, consider a user graph used in a recommendation system, where node labels typically represent user preferences for products. Due to a lack of interest, some users might submit arbitrary or duplicated feedback, resulting in noisy node labels. These noisy labels are demonstrated to significantly impair the performance of GNNs \cite{dai2021nrgnn}. 

The problem of \emph{label noise} has been extensively discussed in the context of \textit{independently and identically distributed (i.i.d.)} data such as images, and a commonly adopted solution is the \emph{label refinement strategy} \cite{li2020dividemix, diao2025mitigating}. This strategy typically entails a two-stage iterative process: updating the model on selected high-confidence labels; refining other labels based on model predictions. While such a strategy could mitigate label noise on \textit{i.i.d.} data, it cannot provide comparable benefits for semi-supervised node classification with label noise (NCLN). It arises because only a limited number of nodes possess label in graph, thus high-confidence labels are even fewer. Therefore, selecting sufficient reliable labels for training becomes more challenging.

To enhance label refinement strategy on graph data, mainstream work frames NCLN as a \emph{reliable labeling} task, constructing reliable pseudo-labels for unlabeled nodes while refining existing labels \cite{qian2023robust}. To more accurately evaluate the reliability, these methods rely on aggregating label information from similar nodes. However, they have limitations in both efficiency and effectiveness. First, for $N$ nodes with feature dimension $d$, the pairwise similarity computation introduced to seek similar nodes increases the algorithm's time complexity to $O(d^2N^2)$, making these methods inefficient. Second, when the noise level increases, simply encouraging the label aggregation could be problematic: a node might be similar to many nodes with inaccurate labels; directly using these incorrect signals actually introduces further noise, and eventually leads to unreliable labeling, as illustrated in Fig. \ref{fig:frameintro} (a). This issue can be observed in Fig. \ref{fig:frameintro} (b), where we calculated the error rates of pseudo-labels: existing works fail to discover real labels as the noise rate increases. These erroneous labels, mistakenly assumed to be correct, could further propagate and accumulate as the iteration proceeds.

To address these problems, we introduce the Label Ensemble Graph Neural Network (LEGNN) method, a lower-complexity method for robust GNNs training against label noise. LEGNN reframes NCLN as a label ensemble-based Partial Label Learning (PLL) task, where it learns a classifier from a set of potentially correct labels for each node \cite{yan2020partial}. With this idea, LEGNN aims to gather informative multiple labels (Fig. \ref{fig:frameintro} (c)) instead of constructing a single reliable label, avoiding the high-complexity similarity computations and mistaken assumption of erroneous labels as correct. Specifically, we conduct two main steps: bootstrapping neighboring context with random masks, and employing the PLL strategy for model training. In the bootstrap step, we apply random neighbor masks for each node and gather the resulting predicted labels as a high-probability label set. This is crucial because inaccurately labeled neighbors can significantly influence the prediction of a target node's label. By adopting these random masks, we reduce dependency of the target node on any specific neighbor, thus diversifying the bootstrapped label set. In the model training step, we utilize a partial label learning (PLL) based strategy to aggregate the bootstrapped label information to guide the model. In addition, to further counteract potential noise from the bootstrapped high-probability label set, we symmetrically gather the low-probability label set with the smallest prediction probability to provide negative supervision.

Our main contributions can be summarized as follows:
\begin{itemize}[leftmargin=*, itemsep=0pt, topsep=0pt]
    \item We investigate the challenge of robust GNN training against label noise in semi-supervised node classification. Prior research has inadequately addressed this issue, with lower efficiency and severe performance degradation in high-noise scenarios.
    \item We are the first to frame the problem of noisy-label node classification as partial label learning task, and we propose a novel framework with label ensemble, LEGNN. Compared to reliable labeling methods, our ensemble design leverages partial label learning with bootstrapped samples to guide the model training, avoiding the reliance on quadratic complexity similarity computations for reliability assessment and the accumulation of errors when such assessment is inaccurate.
    \item Extensive experiments demonstrate the effectiveness and efficiency of our approach. It shows less performance degradation under high noise rates and exhibits good scalability on datasets with over one hundred thousand nodes and one million edges.
\end{itemize}

\section{Related Work}
In this section, we review existing approaches to addressing label noise in \textit{i.i.d.} and graph-structured data.

\subsection{Deep Learning with Label Noise}
As the size of deep neural networks (DNNs) grows, DNNs can fit almost any data. Unfortunately, this could also make the model memorize label noise, thus affecting the model robustness. Extensive efforts have been made to study this issue when learning with \textit{i.i.d.} data. 
These studies can be divided into two categories: loss-based methods \cite{ghosh2017robust} and data-based methods \cite{han2019deep}. Loss-based methods typically mitigate the influence of noisy labels by modifying the objective function. For instance, \cite{patrini2017making} proposed loss correction matrices to adjust prediction probabilities, while others advocated for robust loss designs such as the generalized cross entropy \cite{zhang2018generalized}, passive loss \cite{wang2025joint} or peer loss \cite{liu2020peer} to reduce sensitivity to outliers. In contrast, data-based methods aim to prioritize samples with reliable labels. Prominent strategies in this domain include reweighting \cite{liu2015classification}, which assigns lower weights to likely noisy samples, label correction \cite{han2019deep, xu2025noisy}, which attempts to revise incorrect labels, and sample selection, which retains reliable samples based on specific filtering criteria, including small loss \cite{jiang2018mentornet}, eigenvector-based analysis \cite{cordeiro2025anne}, and contrastive paradigms \cite{zhang2025psscl}. While the loss-based and data-based methods introduce advantages in handling label noise on \textit{i.i.d.} data, prior works have shown that both strategies exhibit limitations when dealing with graph data \cite{dai2021nrgnn}. For instance, loss-based methods usually assume that the noise is random or bounded, which ignores the dependency and propagation of noise along the graph structure. Data-based methods usually require the reliability or consistency of labels, while ignoring the sparsity and imbalance of labels in graph data. Therefore, directly transferring these techniques to graphs is often ineffective.
\subsection{Node Classification with Label Noise}
GNNs have shown superior performance in graph learning.
Unlike \textit{i.i.d.} data, nodes in graph-structured data are interdependent and correlated. Correspondingly, GNNs make node classifications by aggregating information from neighboring nodes.
This makes GNNs more vulnerable to label noise, as the noise can be propagated and amplified via the neighbor aggregation mechanism, thus affecting the model training, especially in the semi-supervised learning setting with only a few labeled nodes \cite{dai2021nrgnn}. 

To address NCLN, mainstream approaches have framed the problem as a reliable labeling task. 
The key principle behind them involves iteratively selecting labels to train the model and updating node labels based on the model's predictions, ultimately obtaining reliable label assignments. Typically, these methods incorporate supplementary robust modules, such as data augmentation \cite{dai2021nrgnn}, contrastive learning \cite{yuan2023learning}, joint learning \cite{qian2023robust}, peer-informed prototypical sample filtering \cite{li2025leveraging}, and link prediction \cite{in2026training}, to aid in the reliable labeling process. For example, NRGNN \cite{dai2021nrgnn} introduces a node distance-based edge predictor to enrich the neighborhood signals. RTGNN \cite{qian2023robust} builds upon this by incorporating sample filtering and joint learning. ProCon \cite{li2025leveraging} leverages peer-informed prototypical consistency with class prototypes to identify reliable nodes. ICGNN \cite{ju2026identifying} identifies noisy labels through influence contradiction. EvGNN \cite{yi2026evidence} exploits evidence-based uncertainty estimation to obtain reliable supervision. MILAB \cite{liu2026mitigating} further enhances robustness by introducing information bottleneck-guided aggregation and consistency regularization. These used robust modules have achieved good results in handling noise by promoting message propagation between similar nodes. However, as the noise level increases, expanding propagation is more likely to expose a node to its noisy similar nodes. In addition, the introduction of complex robust modules significantly increases the computational complexity of the method.

The method proposed in this paper has a fundamental methodological difference from the above methods. We study a new framework where the problem is formulated as a partial label learning problem, and we use label ensemble instead of reliable labeling strategy to aggregate multiple neighbor contexts, which can reduce the neighbor noise.

\section{Formulation and Preliminaries}
\subsection{Problem Formulation}
We denote a graph as $\mathcal{G} = (\mathcal{V}, \mathcal{E})$, where $\mathcal{V}=\{v_i\}_{i=1}^N$ represents the set of $N$ nodes, and $\mathcal{E} \subseteq \mathcal{V} \times \mathcal{V}$ signifies the edge set. $\mathcal{X} = \{\bm{x}_i\}_{i=1}^N$ constitutes a set of node features, where $\bm{x}_i$ is the feature of node $v_i$. In the context of NCLN, typically, we are provided with $M$ labeled nodes $\mathcal{V}^L=\{v^l_i\}_{i=1}^M$ and their noisy labels $\tilde{\mathcal{Y}}^L = \{\tilde{y}^l_i\}_{i=1}^M$, where $\tilde{y}^l_i$ is the noisy label for $v^l_i$. The objective of our problem is to train a robust GNN node classifier $f_{\theta}(\cdot)$ based on $\mathcal{G}$, $\mathcal{X}$, and $\tilde{\mathcal{Y}}^L$ to predict the true labels $\mathcal{Y}^U$ for unlabeled nodes $\mathcal{V}^U = \mathcal{V}\backslash \mathcal{V}^L$, i.e.,
\begin{equation}
    f_{\theta}(\mathcal{G}, \mathcal{X}, \tilde{\mathcal{Y}}^L) \rightarrow \mathcal{Y}^U .
\end{equation}

\subsection{GNN Basic Architecture}
GNNs have found widespread applications in graph node classification tasks. Typically, GNNs employ a message-passing mechanism to embed the graph structure into node-level representations. Specifically, each layer of the GNN employs an aggregate-then-combine operation to update the representation of each node by integrating the representations of its neighboring nodes. In general, the update of the representation $\bm{h}_i^{(k)}$ for node $v_i$ in the $k$-th layer of GNNs can be formalized as follows:
\begin{equation}
    \begin{aligned}
        \bm{a}_i^{(k)}&=\operatorname{AGGREGATE}^{(k)}(\{\bm{h}^{(k-1)}_j:j \in \mathcal{N}_i\}),\\
        \bm{h}_i^{(k)}&=\operatorname{COMBINE}^{(k)}(\bm{h}^{(k-1)}_i,\bm{a}_i^{(k)}),
    \end{aligned}
\end{equation}
where $\operatorname{AGGREGATE}^{(k)}$ and $\operatorname{COMBINE}^{(k)}$ represent the aggregation and combination operations in the $k$-th layer, respectively, and $\mathcal{N}_i$ represents the index set of neighboring nodes for $v_i$. Graph Convolutional Network (GCN) \cite{kipf2017semi}, as a mainstream GNN model, is used in this paper as the backbone model, whose computation of the representation matrix $H^{(k)}$ for the $k$-th layer can be expressed as follows:
\begin{equation}
    H^{(k)}=\sigma(\tilde{A}H^{(k-1)}W^{(k-1)}),
\end{equation}
where $\sigma$ denotes the activation function, $W^{(k-1)}$ represents the weight matrix of $k-1$ layer. $\tilde{A} = \hat{D}^{-1/2}\hat{A}\hat{D}^{-1/2}$ signifies the normalized adjacency matrix, where $\hat{A}=A+I$ and $\hat{D}$ is the diagnal matrix satisfying \(\hat{D}_{ii} = \sum_{j}\hat{A}_{ij}\), \(I\) is the identity matrix, and \(A \in \mathbb{R}^{N \times N}\) represents the adjacency matrix. If nodes \(v_i\) and \(v_j\) are connected, \(A_{ij} = 1\), otherwise, \(A_{ij} = 0\).

\subsection{PLL with Weighted Loss}
In the partial label learning (PLL) task, each instance typically has a set of potentially correct labels, referred to as partial labels. The objective of PLL is to utilize the partial labels to train a classifier that assigns the ground-truth labels.
Loss weighting \cite{feng2020provably} is a commonly used strategy in PLL: given a partially labeled sample $(\bm{x},\mathcal{Y}_{\bm{x}})$ and a classification model $f(\cdot)$, the weighted loss on this sample can be expressed as follows:
\begin{equation}
\ell(\bm{x},\mathcal{Y}_{\bm{x}})=\sum_{y \in \mathcal{Y}_{\bm{x}}} w_y\ell_c(f(\bm{x}),y)
\end{equation}
where $\mathcal{Y}_{\bm{x}}$ represents the partial labels for $\bm{x}$, $w_y$ denotes the weight of label $y$ in $\mathcal{Y}_{\bm{x}}$ concerning the loss, and $\ell_c$ is a classification loss (e.g., the cross entropy loss).

\section{Method}
\begin{figure*}[htb]
    \centering
    \includegraphics[width=0.99\linewidth]{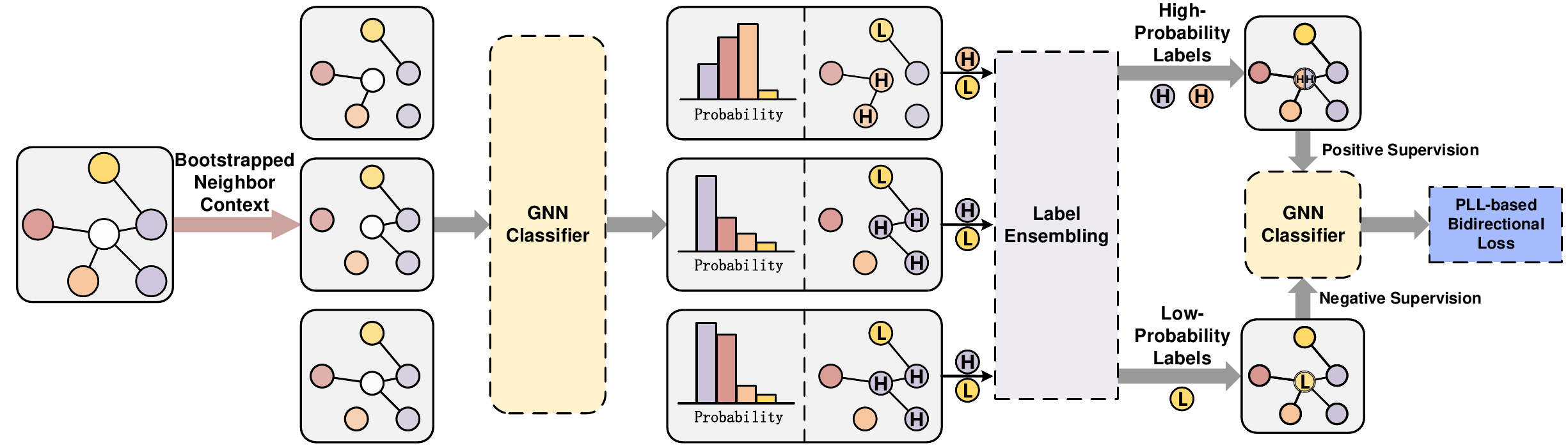} 
    \caption{The overall framework of LEGNN. Nodes are colored differently to indicate their distinct labels, with white representing target nodes to be labeled. LEGNN first generates a masked graph by bootstrapping the neighborhood context. It then infers label probabilities on the masked graph to identify high- and low-probability label sets for the target nodes. Here, the nodes, whose labels match the label with the highest predicted probability for the target node, are masked as ``H'', while the node with the lowest predicted probability for the target node marked as ``L''. Finally, robust model training is performed using the PLL-based strategy with these two label sets.}
    \label{fig:framework}
\end{figure*}

In this section, we introduce the proposed LEGNN framework. Essentially, LEGNN aims to address the potential loss of valuable label information caused by aggressive pseudo-label updates. It achieves this by shifting from reliable labeling to the symmetric ensembling of high-probability and low-probability labels. Additionally, it employs neighbor masking during label ensemble to construct distinct neighboring contexts, thereby averting the sustained influence of specific neighbors. Specifically, LEGNN iteratively performs symmetric ensembling and model training. During symmetric ensembling, we randomly mask edges to generate a set of masked graphs with diverse neighboring contexts. Subsequently, the model serves as an annotator to label these graphs symmetrically, yielding a set of high-probability and low-probability multi-labels. Following this, during model training, a weighted bidirectional loss is introduced to update the model using high-probability and low-probability multi-labels. The overall framework of LEGNN is illustrated in Fig. \ref{fig:framework}.

\subsection{Bootstrapped Neighbor Context}
In graph data, it is commonly assumed that neighboring nodes connected by edges often share the same label \cite{mcpherson2001birds, dai2021nrgnn}. In the context of semi-supervised node classification tasks, this characteristic enables GNNs to utilize message-passing mechanisms, using labeled nodes to make more reliable predictions for the remaining unlabeled nodes. However, in the presence of label noise, GNNs might erroneously assume that nodes connected to these noisy-labeled nodes share the same label. In other words, noisy-labeled nodes can perpetuate confusion in the label estimates of their neighbors, leading to degraded performance of reliable labeling strategy.

\smallskip\noindent\textbf{Bootstrapping via Random Mask} To address this issue, instead of directly constructing a single label as in reliable labeling methods, we propose to first bootstrap the neighbor context by randomly masking a certain ratio of neighbors multiple times. 
Specifically, for each node, we randomly select a proportion $K$ of its neighbors to mask through edge masking, and we conduct this random mask operation for $M_e$ times with replacement. Through this strategy, we generate a set of masked graphs denoted as $\mathcal{G}^m=\{\mathcal{G}_i^m\}_{i=1}^{M_e}$, where $\mathcal{G}_i^m$ represents the masked graph generated in the $i$-th trial. Correspondingly, the set of adjacency matrices of these masked graphs is denoted as $\mathcal{A}^m=\{A_i^m\}_{i=1}^{M_e}$, where $A_i^m$ represents the adjacency matrix associated with $\mathcal{G}_i^m$.

\smallskip\noindent\textbf{Discussion} We view these masked graphs as bootstrapped samples, each generated using an independently sampled random mask. Such a design encourages neighborhood diversity in the resulting graphs, thus is less influenced by the noise on any specific neighboring node. To illustrate the potential benefits of this design, we consider an idealized majority-voting setting with independently corrupted neighbor labels. Suppose a target node with $p$ neighboring nodes whose true labels match the target label and are independently corrupted with probability $\alpha$. Directly generating refined labels by averaging neighbors is exposed to a noise ratio of $\alpha$; on the contrary, if we aggregate the bootstrapped neighbor context and then vote, the error rate is upper-bounded by $\sum_{j=p/2}^{p} \binom{p}{j} \alpha^j (1-\alpha)^{p-j}$, where ties are counted as errors. This bound can be smaller than $\alpha$ when $\alpha<0.5$ and $p$ is sufficiently large. Therefore, bootstrapping neighbor context could reduce noise in the aggregation under a mild assumption that the neighbor information is not fully ruined by noise (i.e., the error rate is less than 0.5). This simplified analysis illustrates the potential benefit of aggregating neighborhood information, although predictions from different masked graphs may not be independent.

\subsection{Symmetric Label Ensemble}
\label{sec:ensemble}
Upon obtaining the masked graphs, LEGNN further ensembles the labeling results from multiple bootstrapped neighbor samples: we first employ the current GNN model to infer the node label given its masked neighbor context; and then we aggregate the labeling results. In this process, a symmetric ensemble strategy is used to gather both high-probability and low-probability labels for all nodes, forming two multi-label sets. 

\smallskip\noindent\textbf{Inferring Label Probability}
Given a graph $\mathcal{G}$ with node features $\mathcal{X}$ and adjacency matrix $A$, the label probability for all nodes can be inferred using the backbone GNN, represented as follows:
\begin{equation}
    Z=f_\theta(A,\mathcal{X}),
\end{equation}
where $f_\theta$ denotes the GNN model, $\theta$ represents its parameters; $Z\in \mathbb{R}^{N\times C}$ represents the model’s output matrix, where $C$ is the number of target classes, and $Z_{ij}$ represents the probability that the model predicts node $v_i$ as class $j$. Empirically, we initialize $\theta$ by minimizing the cross-entropy loss on the noisy data, and then employ $f_\theta$ as the backbone model to generate the label probability for each node given its neighbors.

\smallskip\noindent\textbf{High-Probability Label Ensemble}
For the $i$-th masked graph $\mathcal{G}_i^m$, the computation process of predicting the high-probability labels for its nodes can be formalized as:
\begin{equation}
    \begin{aligned}
        Z^i&=f_\theta(A_i^m,\mathcal{X}),\\
        \hat{\mathcal{Y}}^p_i&=\{\hat{y}^i_k|\hat{y}^i_k=\argmax_{j \in \{1,\cdots,C\}}Z^i_{kj}\}_{k=1}^N,
    \end{aligned}
    \label{eq:yp1}
\end{equation}
where $Z^i$ represents the backbone model's predictive probability on $\mathcal{G}_i^m$, $\hat{\mathcal{Y}}^p_i$ gathers the high-probability labels for all nodes predicted based on the neighbor context in masked graph $\mathcal{G}_i^m$, and $\hat{y}^i_k$ stands for the prediction for node $v_k$ under the $i$-th masked graph. By extending this process to all masked graphs, we can obtain a collection of label sets for all nodes in multiple bootstrapped neighbor contexts, denoted as $\mathcal{Y}^p=\{\hat{\mathcal{Y}}^p_i\}_{i=1}^{M_e}$.

Subsequently, we ensemble these labels to construct the high-probability multi-labels. Concretely, we create a high-probability multi-label matrix $Y^p\in \mathbb{R}^{N\times C}$, where $Y^p_{ij}=1$ indicates that the $j$-th label is a high-probability label for node $v_i$, otherwise, it indicates that it is not a high-probability label. This can be formalized as follows:
\begin{equation}
    Y^p_{ij}=\begin{cases}
        1& \hat{y}^k_i=j,\exists k\in \{1,\cdots,M_e\}\\
        0& \hat{y}^k_i\neq j,\forall k\in \{1,\cdots,M_e\}
    \end{cases}.
    \label{eq:yp2}
\end{equation}
Consequently, the $i$-th row $\boldsymbol{y}_i=Y^p_{i}$ corresponds to the high-probability multi-labels for node $v_i$. This matrix retains predicted labels from different neighbor contexts, avoiding strong misguidance to model where a single erroneous label is considered correct.

\smallskip\noindent\textbf{Low-Probability Label Ensemble}
The resulting high-probability multi-label matrix could also include potentially incorrect labels, due to cognitive bias or improper neighbor masking. Thus, we aim to simultaneously gather auxiliary low-probability labels to provide negative supervision for smoothing the errors in high-probability labels. This process is adopted because, in a general $C$-class ($C \geq 2$) classification task, selecting incorrect labels is easier than selecting the correct one. Consider a model with classification accuracy $\alpha$, tasked with selecting correct and incorrect labels. For selecting an incorrect label, even with a simple strategy that excludes the predicted label and randomly chooses from the remaining ones, the hit rate is $\alpha + (1-\alpha)\frac{C-2}{C-1} \geq \alpha$. Thus, selecting incorrect labels incurs no greater error under this strategy, and as the number of classes increases, this error approaches zero. Consequently, assuming that the model's probability rankings remain informative, using negative supervision from low-probability labels, which are more likely incorrect, helps mitigate the noise effect in the high-probability labels.

Similar to the calculation of high-probability labels, the computation process for low-probability labels can be formalized as:
\begin{equation}
    \hat{\mathcal{Y}}^n_i=\{\bar{y}^i_k|\bar{y}^i_k=\argmin_{j \in \{1,\cdots,C\}}Z^i_{kj}\}_{k=1}^N,
    \label{eq:yn1}
\end{equation}
where $\hat{\mathcal{Y}}^n_i$ represents the set of low-probability labels predicted for all nodes within the contextual information provided by the masked graph $\mathcal{G}_i^m$, and $\bar{y}^i_k$ denotes the predicted incorrect label for node $v_k$ under the $i$-th masked graph. The low-probability multi-label matrix $Y^n\in \mathbb{R}^{N\times C}$ can be formally expressed as
\begin{equation}
    Y^n_{ij}=\begin{cases}
        1& \bar{y}^k_i=j,\exists k\in \{1,\cdots,M_e\}\\
        0& \bar{y}^k_i\neq j,\forall k\in \{1,\cdots,M_e\}
    \end{cases}.
    \label{eq:yn2}
\end{equation}
Here, $Y^n_{ij}=1$ indicates that label $j$ is a low-probability label for node $v_i$, and $\boldsymbol{y}^n_i=Y^n_{i}$ represents the gathered low-probability multi-labels for node $v_i$. By ensembling the labeling results from multiple bootstrapped neighbor contexts, we generate high-probability and low-probability multi-labels for each node. This design can prevent noisy-labeled nodes from affecting the estimation of their neighbors. Additionally, symmetric ensemble can provide richer supervised information for model training.

\subsection{PLL with Weighted Bidirectional Loss} \label{sec:pll_loss}
To effectively utilize the collected high- and low-probability multi-label sets, we design a modular and efficiency-oriented training objective tailored for learning under label ambiguity. While our label construction process is fully decoupled from the subsequent training procedure—allowing flexible integration with any PLL method—we adhere to two key design goals: (1) to extract informative supervision from ambiguous multi-labels while mitigating the noise in high-probability predictions; and (2) to ensure training efficiency by avoiding complex robust modules. 

Benefiting from the quality of our label ensemble, we find that even a simple PLL-based weighted loss~\cite{feng2020provably}—with only linear cost in $O(CN)$—can yield competitive performance. Therefore, we directly instantiate our loss module as a bidirectional extension of this weighted loss, which retains high efficiency while further enabling the use of low-probability multi-labels to counteract noise from the high-probability set in a symmetric and principled manner. 

Specifically, given GNN outputs $Z=f_\theta(A,X)$, the weighted high-probability label loss $\mathcal{L}^p$ can be expressed as follows:
\begin{equation}
    \mathcal{L}^p(Z,Y^p)=\frac{1}{N}\sum_{v_i\in \mathcal{V}}\sum_{j=1}^{C}w(Z_{i},Y^p_{ij})\ell(Z_{i},j),
    \label{eq:positive}
\end{equation}
where, $Z_{i}$ represents the $i$-th row of $Z$, corresponding to the predicted output of the network for node $v_i$, and $\ell(Z_{i},j)$ denotes the employed classification loss. The weighting function $w(\cdot)$ is formulated as 
$w(Z_{i},Y^p_{ij})={(Z_{ij}Y^p_{ij})}{/(\sum_{k=1}^{C}Z_{ik}Y^p_{ik})}$, where $w(Z_{i},Y^p_{ij})$ denotes the weight based on normalized network prediction confidence which reflects the impact of the classification loss between node $v_i$ and label $Y^p_{ij}$ on model training. Intuitively, a higher weight suggests a higher probability of label $Y^p_{ij}$ is the correct label for node $v_i$. In implementation, we detach the weights from the computational graph to prevent gradient propagation through them.

Correspondingly, in order to counteract noise from the high-probability label set, we symmetrically employ the following loss as the low-probability label loss: $\mathcal{L}^n(Z,Y^n)=\mathcal{L}^p(O-Z,Y^n)$, 
where $O\in \mathbb{R}^{N \times C}$ denotes a matrix with all elements being 1, i.e., for any matrix element, $O_{ij}=1$. Intuitively, this loss function can generate negative supervised signals to counteract signals that are mistakenly considered correct. Ultimately, the overall weighted bidirectional loss is defined as follows:
\begin{equation}
    \mathcal{L}(Z,Y^p,Y^n)=\mathcal{L}^p(Z,Y^p)+\mathcal{L}^n(Z,Y^n).
    \label{eq:loss}
\end{equation}
This loss serves as the default instantiation in LEGNN due to its efficiency. Nevertheless, it can be easily replaced with any other PLL loss $\mathcal{L}^p$ to trade off performance and cost as needed. As an example, we combine our label ensemble with a more sophisticated PLL method—PLD~\cite{patel2024progressive}—to form the LEPLD variant, whose experimental evaluation is presented in Section~\ref{sec:node_cls}, with implementation details provided in the Supplementary Material Section 1.

\subsection{LEGNN Training Framework}
The overall training process of LEGNN is presented in Algorithm \ref{alg:algorithm}, which can be divided into two phases: initialization and model training. In the initialization phase, LEGNN initializes a model $f_\theta$ using noisy data and a classification loss. In the subsequent model training phase, LEGNN iteratively conducts bootstrapping of neighboring contexts and model training. Specifically, for bootstrapping neighboring contexts, LEGNN applies multiple edge masks to generate masked graphs first. These graphs are then labeled with the model $f_\theta$, and the labels are gathered. For model training, LEGNN updates the parameters of $f_\theta$ using the gathered labels and the weighted bidirectional loss. A running example appears in the Supplementary Material Section 2.

\begin{algorithm}[htb]
    \caption{LEGNN training framework}
    \label{alg:algorithm}
    \begin{algorithmic}[1]
        \REQUIRE Adjacency matrix $A$, feature matrix $X$, initial model $Z=f_\theta(A,X)$, training set with noisy labels $\tilde {\mathcal{Y}}$, validation set with noisy labels $\tilde {\mathcal{Y}}^v$, mask rate $K$, mask times $M_e$, classification loss $\ell$.\\
        \ENSURE Classifier model $f$.
        \STATE Train $f_\theta$ using loss $\ell$ and noisy labels $\tilde {\mathcal{Y}}$ along with corresponding node features.
        \STATE Construct masked adjacency matrix $\mathcal{A}$ based on $K$ and $M_e$.
        \STATE Employ $f_\theta$ to symmetrically predict the masked graph, and create high-probability and low-probability multi-label matrices $Y^P$ and $Y^n$ as per Eq. (\ref{eq:yp1}-\ref{eq:yn2}).
        \WHILE{not convergence}
            \STATE Compute loss using Eq. \eqref{eq:loss}, then update $\theta$.
            \IF {validation classification accuracy declined}
            \STATE Recalculate the weight.
            \STATE Perform label gathering again following step 2 and 3.
            \ENDIF
        \ENDWHILE
        \STATE \textbf{return} Classifier $f_\theta$ with optimized $\theta$.
    \end{algorithmic}
\end{algorithm}

\subsection{Time Complexity Analysis}
\label{sec:complexity}
In this section, we analyze the additional training complexity of LEGNN. Let $M_e$, $C$, $|E|$, $N$, $d$, $T$, and $R$ denote the number of masks per label generation, classes, edges, nodes, hidden dimension, training epochs, and label generation rounds, respectively. LEGNN introduces an additional complexity of $comp=O(RM_e|E|+RM_eF(N,|E|)+RM_eCN+2TCN)=O(RM_e(1+Ld)|E|+[RM_eLd^2+(RM_e+2T)C]N)$ on top of backbone training, where $F(N,|E|)=O(Ld^2N+Ld|E|)$ represents the forward complexity of an $L$-layer GCN under a uniform-width approximation \cite{guo2023hierarchical}. Specifically, these terms account for masked graph construction, masked prediction, label ensemble, and weighted bidirectional loss computation, respectively. This shows that LEGNN's additional complexity depends on graph connectivity. For sparse graphs with $|E|\leq N\log N$, we have $[RM_eLd^2+(RM_e+2T)C]N\leq RM_e(1+Ld)|E|+[RM_eLd^2+(RM_e+2T)C]N\leq RM_e(1+Ld)N\log N+[RM_eLd^2+(RM_e+2T)C]N$, with the complexity level ranging from $O(N)$ to $O(N\log N)$ when the remaining parameters are fixed. For denser graphs with $N\log N<|E|\leq N^2$, we obtain $RM_e(1+Ld)N\log N+[RM_eLd^2+(RM_e+2T)C]N<RM_e(1+Ld)|E|+[RM_eLd^2+(RM_e+2T)C]N\leq RM_e(1+Ld)N^2+[RM_eLd^2+(RM_e+2T)C]N$, with the complexity level ranging from $O(N\log N)$ to $O(N^2)$ as the edge count increases. In other words, on the sparse graphs commonly encountered in practice, LEGNN's additional complexity has a lower growth order than the $O(N^2)$ cost of reliable labeling methods that require all-pairs node comparisons.

\section{Experiments} \label{sec:experiments}
In this section, we conduct experiments to address the following research questions:
\begin{itemize}[leftmargin=*, itemsep=0pt, topsep=0pt]
    \item[] \textbf{RQ1}: Does the proposed approach exhibit robustness across various noise settings and label rates?
    \item[] \textbf{RQ2}: What is the scalability of the proposed method?
    \item[] \textbf{RQ3}: Does the proposed method entail lower computational costs compared to methods utilizing intricate modules?
    \item[] \textbf{RQ4}: Constructing such a label set may introduce more erroneous labels (or more noise). However, why doesn't this lead to a decrease in performance?
    \item[] \textbf{RQ5}: How do components and parameters affect LEGNN?
\end{itemize}
\subsection{Dataset Details} %
\noindent\textbf{Dataset} We present experiments on six node classification benchmarks (Cora, Citeseer, Pubmed \cite{sen2008collective}, IGB \cite{khatua2023igb}, OGBN-Arxiv \cite{hu2020open}, Actor \cite{pei2020geom} ) and one real-world noisy dataset (Clothing1m \cite{xiao2015learning}). Among them, Actor is a heterophilic dataset, which we include to explore the generalizability of our method despite its non-target design. For the first six datasets with clean original labels, we inject three widely used noise types: \emph{symmetric flip} \cite{patrini2017making}, \emph{pair flip} \cite{han2018co}, and \emph{instance-dependent} \cite{xia2020part} noise. Dataset statistics are summarized in Table \ref{tab:datasets}. Following RTGNN~\cite{qian2023robust}, we use 5\% of the nodes for training and 15\% for validation on Cora, Citeseer, IGB, and Actor; 1\% for training and 19\% for validation on Pubmed; and the default split for OGBN-Arxiv. For Clothing1m, following \cite{xia2023gnn}, we construct a nearest-neighbor graph (Clothing1m-graph, C1mG) by sampling about 15k examples to enable graph-based learning with 5k for training. In all experiments, only noisy labels are available during training and validation. Specifically, noisy training labels are used for model training, noisy validation labels for training monitoring and model selection, and clean test labels exclusively for final evaluation.

\begin{table*}[htb]
    \centering
    \resizebox{0.99\linewidth}{!}{
    \begin{tabular}{lllllllllllllllllll}
        \toprule
        \multirow{4}{*}{Dataset} & \multirow{4}{*}{Nodes} & \multirow{4}{*}{Edges} & \multirow{4}{*}{Features} & \multirow{4}{*}{$C$} & \multicolumn{14}{c}{Actual Noise Rate(\%)} \\
        \cmidrule(lr){6-19}
        & & & & & \multicolumn{5}{c}{Sym} & \multicolumn{4}{c}{Pair} & \multicolumn{5}{c}{IDN} \\
        \cmidrule(lr){6-10}\cmidrule(lr){11-14}\cmidrule(lr){15-19}
        & & & & & 10\% & 20\% & 30\% & 40\% & 50\% & 10\% & 20\% & 30\% & 40\% & 10\% & 20\% & 30\% & 40\% & 50\%\\
        \midrule
        Cora & 2,485 & 5,068 & 1,433 & 7 & 8.94 & 22.58 & 32.52 & 42.14 & 53.66 & 12.90 & 18.70 & 27.82 & 36.59 & 8.13 & 19.96 & 27.64 & 39.92 & 50.41 \\ 
        Citeseer & 2,110 & 3,668 & 3,703 & 6 & 5.71 & 24.70 & 33.33 & 41.57 & 56.19 & 12.11 & 20.00 & 28.50 & 35.24 & 7.62 & 20.95 & 28.57 & 36.19 & 48.57 \\ 
        Pubmed & 19,717 & 44,338 & 500 & 3 & 11.68 & 19.15 & 27.41 & 39.23 & 51.27 & 10.45 & 17.26 & 29.52 & 39.59 & 11.17 & 18.27 & 31.47 & 44.67 & 51.78 \\ 
        IGB & 100,000 & 547,416 & 1,024 & 19 & 9.98 & 20.10 & 29.49 & 39.72 & 50.23 & 10.08 & 19.68 & 29.51 & 38.91 & 10.24 & 20.12 & 29.35 & 39.71 & 49.83 \\ 
        OGBN & 169,343 & 1,166,243 & 128 & 40 & 10.13 & 20.07 & 29.96 & 40.14 & 49.88 & 10.09 & 20.12 & 30.10 & 40.07 & 10.04 & 19.96 & 30.08 & 40.07 & 49.94 \\ 
        Actor & 7,600 & 33,544 & 931 & 5 & 11.08 & 21.46 & 30.61 & 41.21 & 52.24 & 11.65 & 19.79 & 31.07 & 37.73 & 10.55 & 22.96 & 29.02 & 39.57 & 46.17 \\ 
        C1mG & 15,524 & 77,620 & 1,000 & 14 & \multicolumn{14}{c}{-} \\ 
        \bottomrule
    \end{tabular}
    }
    \caption{Statistics of datasets. $C$ denotes number of classes.}
    \label{tab:datasets}
\end{table*}

\smallskip\noindent\textbf{Synthetic Noise Generation} The three types of label noise simulate distinct annotator error patterns: (a) symmetric flip noise models random errors across any class \cite{patrini2017making}, (b) pair flip noise captures errors confined to semantically similar classes \cite{han2018co}, and (c) instance-dependent noise (IDN) \cite{xia2020part} reflects annotator inconsistencies due to instance-specific factors such as ambiguous node features. For symmetric and pair flip noise, we randomly flip the labels according to a transition matrix $T$, where $T_{ij}$ represents the probability of the $i$-th class label being assigned as the $j$-th class. Given a noise rate $\tau$, the respective transition matrices $T_{\text{sym}}, T_{\text{pair}} \in \mathbb{R}^{C\times C}$ are defined as follows:
\begin{equation}
\begin{aligned}
    T_{symmetric}&=\left[\begin{array}{cccc}1-\tau & \frac{\tau}{C-1} & \cdots & \frac{\tau}{C-1} \\ \frac{\tau}{C-1} & 1-\tau & \cdots & \frac{\tau}{C-1} \\ \vdots & \vdots & \ddots & \vdots \\ \frac{\tau}{C-1} & \cdots & \frac{\tau}{C-1} & 1-\tau\end{array}\right]_{C \times C},
\end{aligned}
\end{equation}
\begin{equation}
\begin{aligned}
    T_{pair}&=\left[\begin{array}{cccc}1-\tau & 0 & \cdots & \tau \\ \tau & 1-\tau & \cdots & 0 \\ \vdots & \vdots & \ddots & \vdots \\ 0 & \cdots & \tau & 1-\tau\end{array}\right]_{C \times C}.
\end{aligned}
\end{equation}

For IDN, the instance-dependent label noise generation process synthesizes noisy labels by sampling instance-specific flip probability from a feature based truncated normal distribution, following \cite{xia2020part}. We additionally provide a concrete example of the synthetic noise in the Supplementary Material Section 3.1.

\smallskip\noindent\textbf{Construction of the C1mG} For Clothing1m, we follow \cite{xia2023gnn} to sample the training data and construct a nearest-neighbor graph to facilitate graph-based learning. Specifically, we randomly sampled 4,998 instances from the noisy training data of Clothing1m to form a pseudo-balanced subset as training nodes, while using all test instances from Clothing1m as test nodes. Then we employed a pre-trained ResNet-50 \cite{he2016deep} on ImageNet to extract features and utilized a KNN (k=5) algorithm to build a nearest-neighbor C1mG.

\subsection{Implementation and Baseline}
\smallskip\noindent\textbf{Implementation Details} We utilize a dual-layer GCN with a hidden dimension of 64 as the backbone network. Stochastic Gradient Descent (SGD) is used as the optimizer for the Cora, Citeseer, Pubmed, OGBN-Arxiv and Actor benchmarks, while Adam is employed as the optimizer for the IGB. All training was conducted for 200 epochs, with a fixed momentum of 0.9 and a dropout rate of 0.5. Furthermore, guided by the validation set accuracy, we select learning rates from $\{0.001, 0.005, \dots, 4\}$, weight decay from $\{1 \times 10^{-6}, 5 \times 10^{-6}, \dots, 0.1\}$, the number of mask times from $\{5, 10, \dots, 30\}$, and the mask rate from $\{0.1, \dots, 0.9\}$. In the experiments, we set the mask rate of C1mG to $0.2$ and the mask times to $3$. Due to its higher number of edges, the mask rate for OGBN-Arxiv was set to $0.8$ with $20$ mask times. For the remaining datasets, the mask rate was set to $0.5$ and the mask times to $10$. All experiments were conducted on the RTX 4090 GPUs. Our code is available at https://github.com/RyanZhaoIc/LEGNN.git.

\smallskip\noindent\textbf{Baselines} 
We compare LEGNN with GNN-based approaches for noisy node classification, including 
\textbf{GCN} \cite{kipf2017semi}, a spectral graph convolutional network; 
\textbf{Self-training} \cite{li2018deeper}, using pseudo labeling to iteratively expand; 
\textbf{Forward} \cite{patrini2017making}, introducing a classifier-consistency loss via a transition matrix; 
\textbf{CP} \cite{zhang2020adversarial}, incorporating community labels obtained by clustering node embeddings into GCN training; 
\textbf{Coteaching+} \cite{yu2019does}, using two networks to mutually select small-loss samples for cooperative training; 
\textbf{NRGNN} \cite{dai2021nrgnn}, linking unlabeled and labeled nodes with pseudo-labels to mitigate noise; \textbf{RTGNN} \cite{qian2023robust}, which extends NRGNN with an explicit noise-governance strategy; and \textbf{ProCon} \cite{li2025leveraging}, which leverages peer-informed prototypical consistency to identify reliable nodes. 
For methods originally designed for \textit{i.i.d.} data, such as Forward and Coteaching+, we follow NRGNN and adopt the GCN backbone for a fair comparison.

\subsection{Experimental Results of Node Classification} 
\label{sec:node_cls}
To address \textbf{RQ1} and \textbf{RQ2}, we compared the classification results of LEGNN and its LEPLD variant (introduced in Section \ref{sec:pll_loss}) with various baselines. Specifically, we evaluated under different noise types, noise rates, and label rates.

\smallskip\noindent\textbf{Experimental Results under Different Noise Settings and Label Rates (\textbf{RQ1})} 
Tables~\ref{tab:acc_small} and \ref{tab:acc_large} report results on standard and large-scale datasets, respectively (complete results for Sym-10\%--50\%, Pair-10\%--40\%, and IDN-10\%--50\% are deferred to Supplementary Material Section 5). Additionally, Table~\ref{tab:Clothing1m} reports results under real-world noise. Our methods (LEGNN/LEPLD) achieve the highest mean accuracy in 37 of 43 cases. Although some performance differences are relatively small compared to the reported standard deviations, larger improvements can be observed under several challenging noise settings. For instance, under high noise settings such as Sym-50\%, LEGNN achieves up to 5.63\% improvement (Citeseer), and it also yields a 3.19\% gain on the real-world  dataset. On sparser graphs such as IGB and Citeseer, the LEPLD variant—enhanced by an edge augmentation strategy from PLD \cite{patel2024progressive}—further improves the performance of LEGNN, highlighting the extensibility of our framework to more advanced PLL strategies. 

\begin{table*}[t]
\centering
\resizebox{0.9\textwidth}{!}{
    \begin{tabular}{llccccccc}
    \toprule
    Dataset & Method & Clean & Sym-20\% & Sym-50\% & Pair-20\% & Pair-40\% & IDN-20\% & IDN-50\% \\
    \midrule
    \multirow{10}{*}{Cora}
    & GCN           & 81.32 ± 1.15 & 71.47 ± 1.66 & 53.05 ± 3.89 & 76.00 ± 1.11 & 58.08 ± 1.51 & 74.88 ± 0.97 & 50.96 ± 4.87 \\
    & Self-training & 78.21 ± 1.53 & 72.52 ± 1.42 & 53.46 ± 3.29 & 76.03 ± 0.97 & 58.56 ± 1.99 & 75.63 ± 0.67 & 53.27 ± 4.46 \\
    & Forward       & 81.50 ± 1.20 & 72.19 ± 0.83 & 52.54 ± 4.66 & 75.20 ± 1.55 & 57.75 ± 1.25 & 75.40 ± 1.21 & 52.81 ± 4.19 \\
    & CP            & 79.97 ± 1.31 & 68.58 ± 1.50 & 47.81 ± 2.11 & 73.69 ± 0.09 & 53.50 ± 1.85 & 72.54 ± 0.89 & 50.64 ± 4.35 \\
    & Coteaching+   & 78.21 ± 1.53 & 67.87 ± 1.91 & 51.07 ± 4.44 & 72.88 ± 2.50 & 52.76 ± 6.17 & 72.20 ± 1.26 & 46.37 ± 2.66 \\
    & NRGNN         & 81.31 ± 0.89 & \textbf{80.02 ± 0.77} & 64.72 ± 2.26 & 79.50 ± 1.85 & 66.06 ± 2.88 & 79.58 ± 1.13 & 62.61 ± 1.81 \\
    & RTGNN         & 79.67 ± 1.17 & 76.82 ± 1.28 & 63.05 ± 3.15 & 77.24 ± 0.91 & 67.51 ± 2.69 & 78.53 ± 0.67 & 58.14 ± 3.67 \\
    & ProCon        & 79.44 ± 1.25 & 76.34 ± 3.48 & 66.96 ± 2.80 & 77.51 ± 1.73 & 63.97 ± 4.42 & 73.80 ± 0.99 & 67.90 ± 1.88 \\
    \cmidrule(lr){2-9}
    & LEGNN         & \textbf{81.57 ± 0.77} & 79.95 ± 1.56 & \textbf{67.93 ± 3.81} & \textbf{79.64 ± 1.00} & 67.55 ± 0.64 & \textbf{79.96 ± 2.12} & \textbf{68.98 ± 4.47} \\
    & LEPLD         & 78.27 ± 0.86 & 77.07 ± 0.69 & 61.53 ± 2.66 & 78.90 ± 0.77 & \textbf{75.42 ± 0.86} & 77.80 ± 1.09 & 66.33 ± 3.64 \\
    \midrule
    \multirow{10}{*}{Citeseer}
    & GCN           & 72.09 ± 1.01 & 62.73 ± 1.40 & 46.78 ± 0.70 & 66.37 ± 1.02 & 49.91 ± 0.96 & 64.62 ± 1.53 & 46.54 ± 5.44 \\
    & Self-training & 72.20 ± 1.09 & 63.61 ± 2.21 & 46.88 ± 0.48 & 66.75 ± 0.74 & 50.26 ± 1.09 & 64.74 ± 1.64 & 48.48 ± 4.79 \\
    & Forward       & 72.42 ± 1.02 & 63.58 ± 1.45 & 46.66 ± 0.99 & 67.00 ± 1.33 & 51.64 ± 1.35 & 64.42 ± 2.36 & 47.07 ± 4.30 \\
    & CP            & 71.56 ± 0.79 & 62.31 ± 1.66 & 46.47 ± 3.33 & 65.88 ± 1.80 & 49.92 ± 1.86 & 63.07 ± 2.07 & 47.44 ± 3.79 \\
    & Coteaching+   & 68.98 ± 0.84 & 59.95 ± 2.83 & 44.43 ± 5.35 & 63.42 ± 2.30 & 50.84 ± 4.42 & 61.04 ± 2.71 & 42.01 ± 2.35 \\
    & NRGNN         & 73.54 ± 1.02 & 68.08 ± 2.19 & 55.07 ± 3.72 & 69.08 ± 1.57 & 52.77 ± 3.65 & 72.00 ± 0.42 & 63.54 ± 1.50 \\
    & RTGNN         & 71.00 ± 1.42 & 67.22 ± 0.99 & 54.56 ± 2.43 & 67.77 ± 1.22 & 56.52 ± 2.71 & 67.82 ± 2.16 & 59.50 ± 2.70 \\
    & ProCon        & 73.48 ± 0.65 & 71.64 ± 2.99 & 63.99 ± 4.86 & 71.86 ± 3.06 & 58.08 ± 4.01 & 71.81 ± 0.63 & 69.02 ± 0.60 \\
    \cmidrule(lr){2-9}
    & LEGNN         & 76.28 ± 0.33 & 74.70 ± 0.87 & 69.62 ± 3.69 & 73.38 ± 1.64 & 64.82 ± 2.20 & \textbf{73.59 ± 1.33} & \textbf{69.80 ± 2.28} \\
    & LEPLD         & \textbf{77.88 ± 0.62} & \textbf{77.34 ± 0.40} & \textbf{75.19 ± 1.92} & \textbf{76.74 ± 0.39} & \textbf{69.63 ± 3.06} & 72.92 ± 0.82 & 66.02 ± 1.14 \\
    \midrule
    \multirow{10}{*}{Pubmed}
    & GCN           & 81.17 ± 1.82 & 74.08 ± 0.20 & 57.62 ± 3.06 & 70.50 ± 0.31 & 60.43 ± 3.10 & 78.21 ± 2.03 & 56.74 ± 3.76 \\
    & Self-training & 81.12 ± 1.90 & 74.49 ± 0.11 & 58.51 ± 3.44 & 70.56 ± 0.34 & 60.31 ± 3.66 & 78.65 ± 2.14 & 57.44 ± 3.58 \\
    & Forward       & \textbf{81.27 ± 1.72} & 74.73 ± 0.14 & 57.61 ± 3.20 & 71.59 ± 0.22 & 61.71 ± 4.48 & \textbf{78.81 ± 2.28} & 56.61 ± 2.73 \\
    & CP            & 80.87 ± 1.70 & 74.04 ± 0.32 & 58.55 ± 2.58 & 70.32 ± 0.31 & 55.33 ± 3.98 & 78.35 ± 2.24 & 58.28 ± 2.72 \\
    & Coteaching+   & 80.30 ± 1.91 & 71.30 ± 2.74 & 51.99 ± 7.96 & 68.82 ± 3.57 & 60.21 ± 2.72 & 76.76 ± 3.23 & 55.10 ± 4.61 \\
    & NRGNN         & 69.48 ± 1.10 & 71.61 ± 1.16 & 56.44 ± 1.52 & 66.14 ± 0.91 & 51.81 ± 3.12 & 63.93 ± 2.23 & 60.85 ± 1.00 \\
    & RTGNN         & 73.62 ± 0.57 & 62.86 ± 3.83 & 50.42 ± 3.08 & 58.64 ± 2.47 & 58.66 ± 3.86 & 75.04 ± 0.53 & 62.90 ± 5.18 \\
    & ProCon        & 79.24 ± 1.41 & 75.01 ± 1.20 & 59.17 ± 1.64 & \textbf{73.13 ± 1.58} & 62.84 ± 4.35 & 75.81 ± 0.02 & 61.23 ± 1.29 \\
    \cmidrule(lr){2-9}
    & LEGNN         & 79.88 ± 0.43 & 74.92 ± 0.33 & \textbf{61.28 ± 0.78} & 72.35 ± 0.27 & 61.79 ± 3.22 & 76.12 ± 0.57 & \textbf{63.67 ± 1.40} \\
    & LEPLD         & 78.49 ± 0.30 & \textbf{75.55 ± 0.74} & 59.25 ± 0.67 & 72.62 ± 0.22 & \textbf{67.85 ± 1.19} & 75.20 ± 0.52 & 60.63 ± 1.72 \\
    \midrule
    \multirow{10}{*}{Actor}
    & GCN           & 28.98 ± 0.43 & 25.65 ± 0.28 & 22.78 ± 1.50 & 25.32 ± 0.57 & 24.44 ± 0.35 & 25.40 ± 1.01 & 21.70 ± 1.87 \\
    & Self-training & 28.11 ± 0.26 & 25.78 ± 0.23 & 23.73 ± 1.69 & 25.40 ± 0.47 & 24.32 ± 0.42 & 25.70 ± 0.60 & 21.67 ± 1.57 \\
    & Forward       & 28.24 ± 0.86 & 26.29 ± 0.36 & 24.05 ± 1.34 & 25.49 ± 1.12 & 24.02 ± 2.16 & 26.44 ± 0.91 & 21.52 ± 2.91 \\
    & CP            & 28.22 ± 0.78 & 25.90 ± 0.68 & 23.81 ± 1.96 & 25.32 ± 1.22 & 24.08 ± 0.67 & 25.90 ± 0.96 & 22.15 ± 1.30 \\
    & Coteaching+   & 27.76 ± 0.29 & 24.74 ± 0.43 & 23.85 ± 1.17 & 25.22 ± 0.95 & 23.64 ± 0.79 & 25.59 ± 0.60 & 22.28 ± 1.02 \\
    & NRGNN         & \textbf{29.05 ± 0.19} & 26.92 ± 0.97 & 25.01 ± 0.99 & 27.18 ± 1.16 & 23.98 ± 0.60 & 26.61 ± 0.97 & 22.23 ± 2.44 \\
    & RTGNN         & 28.84 ± 0.14 & 26.38 ± 0.44 & 25.84 ± 0.37 & 26.20 ± 0.15 & 24.52 ± 0.32 & 26.62 ± 0.19 & 22.48 ± 1.66 \\
    & ProCon        & 28.09 ± 0.64 & \textbf{27.84 ± 0.10} & 24.28 ± 0.56 & 26.36 ± 1.09 & 24.81 ± 1.52 & 27.64 ± 0.20 & 24.46 ± 0.07 \\
    \cmidrule(lr){2-9}
    & LEGNN         & 28.32 ± 0.17 & 26.07 ± 0.13 & 26.26 ± 0.60 & 26.35 ± 1.36 & 25.19 ± 1.84 & 26.81 ± 0.33 & 22.51 ± 2.10 \\
    & LEPLD         & 28.55 ± 0.24 & 27.38 ± 0.92 & \textbf{28.05 ± 0.73} & \textbf{29.77 ± 0.89} & \textbf{27.15 ± 0.26} & \textbf{28.63 ± 1.33} & \textbf{29.03 ± 3.83} \\
    \bottomrule
    \end{tabular}
}
\caption{Classification accuracy (\%) $\pm$ standard deviation under different noise settings.}
\label{tab:acc_small}
\end{table*}

\begin{table*}[htb]
\centering
\resizebox{0.95\textwidth}{!}{
    \begin{tabular}{llccccccc}
    \toprule
    Dataset & Method & Clean & Sym-20\% & Sym-50\% & Pair-20\% & Pair-40\% & IDN-20\% & IDN-50\% \\
    \midrule
    \multirow{10}{*}{IGB}
    & GCN           & 67.65 ± 0.27 & 65.94 ± 0.14 & 57.30 ± 4.36 & 66.22 ± 0.40 & 59.57 ± 0.76 & 65.58 ± 0.38 & 63.02 ± 0.41 \\
    & Self-training & 67.63 ± 0.29 & 67.12 ± 0.24 & 64.52 ± 0.40 & 66.29 ± 0.41 & 59.54 ± 0.86 & 67.15 ± 0.38 & 64.51 ± 0.45 \\
    & Forward       & 67.22 ± 0.29 & 66.28 ± 0.47 & 63.82 ± 0.46 & 66.01 ± 0.46 & 59.96 ± 3.51 & 66.54 ± 0.52 & 63.51 ± 0.60 \\
    & CP            & 66.79 ± 0.17 & 64.66 ± 0.42 & 61.31 ± 0.70 & 64.59 ± 0.53 & 56.26 ± 0.90 & 64.69 ± 0.38 & 60.45 ± 0.76 \\
    & Coteaching+   & 66.71 ± 0.25 & 66.30 ± 0.35 & 64.02 ± 0.36 & 65.44 ± 0.29 & 58.53 ± 0.77 & 66.45 ± 0.31 & 63.95 ± 0.68 \\
    & NRGNN         & 63.02 ± 1.38 & 64.79 ± 0.88 & 63.82 ± 0.49 & 63.03 ± 1.15 & 58.11 ± 2.50 & 63.81 ± 0.76 & 63.12 ± 0.29 \\
    & RTGNN         & 68.54 ± 0.45 & 67.67 ± 0.14 & 65.88 ± 0.35 & 65.94 ± 0.14 & 61.08 ± 0.95 & 67.80 ± 0.29 & 67.07 ± 0.52 \\
    & ProCon        & 64.99 ± 0.23 & 63.33 ± 0.58 & 60.59 ± 0.86 & 63.50 ± 0.36 & 58.69 ± 0.98 & 63.29 ± 0.08 & 59.94 ± 0.05 \\
    \cmidrule(lr){2-9}
    & LEGNN         & 68.81 ± 0.11 & 68.60 ± 0.07 & 66.37 ± 0.21 & 66.73 ± 0.17 & 62.54 ± 0.53 & 68.57 ± 0.08 & 67.07 ± 0.11 \\
    & LEPLD         & \textbf{70.16 ± 0.22} & \textbf{70.01 ± 0.13} & \textbf{67.97 ± 0.36} & \textbf{68.50 ± 0.09} & \textbf{66.87 ± 0.06} & \textbf{69.89 ± 0.20} & \textbf{68.72 ± 0.27} \\
    \midrule
    \multirow{10}{*}{OGBN}
    & GCN           & 60.90 ± 1.17 & 46.61 ± 1.40 & 45.43 ± 0.73 & 46.37 ± 1.05 & 44.94 ± 0.81 & 59.66 ± 1.60 & 53.79 ± 1.67 \\
    & Self-training & 60.64 ± 1.15 & 48.09 ± 0.93 & 49.01 ± 0.63 & 46.50 ± 1.22 & 45.29 ± 1.06 & 59.73 ± 1.42 & 54.64 ± 1.50 \\
    & Forward       & 61.32 ± 1.68 & 47.81 ± 1.42 & 45.57 ± 0.73 & 46.10 ± 1.56 & 42.86 ± 2.16 & 58.31 ± 2.04 & 54.22 ± 1.63 \\
    & CP            & 61.12 ± 1.06 & 44.51 ± 0.90 & 41.69 ± 0.95 & 44.87 ± 0.93 & 44.51 ± 1.47 & 60.25 ± 0.97 & 53.07 ± 1.74 \\
    & Coteaching+   & 59.74 ± 1.03 & 46.18 ± 1.81 & 45.85 ± 1.99 & 45.16 ± 2.20 & 43.35 ± 1.01 & 60.62 ± 0.71 & 54.71 ± 0.86 \\
    & NRGNN         & 31.68 ± 4.63 & 35.33 ± 2.87 & 33.06 ± 5.16 & 33.40 ± 2.74 & 42.53 ± 0.04 & 29.09 ± 3.51 & 27.58 ± 0.83 \\
    & RTGNN         & 29.98 ± 0.28 & 31.16 ± 1.48 & 27.44 ± 1.21 & 30.05 ± 0.28 & 31.13 ± 0.07 & 29.80 ± 0.33 & 28.57 ± 0.22 \\
    & ProCon        & 56.24 ± 2.09 & 51.97 ± 2.32 & 43.47 ± 3.83 & 56.41 ± 1.08 & 54.91 ± 0.81 & 54.74 ± 2.26 & 44.60 ± 1.98 \\
    \cmidrule(lr){2-9}
    & LEGNN         & \textbf{64.54 ± 0.21} & \textbf{61.08 ± 0.22} & \textbf{57.48 ± 0.43} & \textbf{59.45 ± 3.72} & \textbf{58.39 ± 0.22} & \textbf{64.17 ± 0.17} & \textbf{62.74 ± 0.19} \\
    & LEPLD         & 46.61 ± 1.39 & 35.59 ± 1.16 & 32.85 ± 12.62 & 43.01 ± 0.56 & 41.84 ± 0.58 & 53.89 ± 10.31 & 34.97 ± 4.48 \\
    \bottomrule
    \end{tabular}
    }
\caption{Classification accuracy (\%) $\pm$ standard deviation on large-scale datasets.}
\label{tab:acc_large}
\end{table*}

\begin{table}[htb]
    \centering
    \resizebox{\linewidth}{!}{
    \begin{tabular}{ccccccccc}
    \toprule
        GCN & Self-training & Forward & CP & Coteaching+ & NRGNN & RTGNN & ProCon & LEGNN\\
    \midrule
        25.07±1.73 & 27.86±1.92 & 26.95±1.31 & 25.27±1.64 & 25.65±0.98 & 30.70±0.92 & 32.72±0.86 & 36.41 ± 0.36 & \textbf{39.60±1.14}\\
    \bottomrule
    \end{tabular}
    }
    \caption{The average classification accuracy and standard deviation (\%) across five trials for C1mG.}
    \label{tab:Clothing1m}
\end{table}

\begin{figure}[htb]
    \centering
    \begin{minipage}{0.4\linewidth}
        \centering
        \includegraphics[width=0.99\columnwidth]{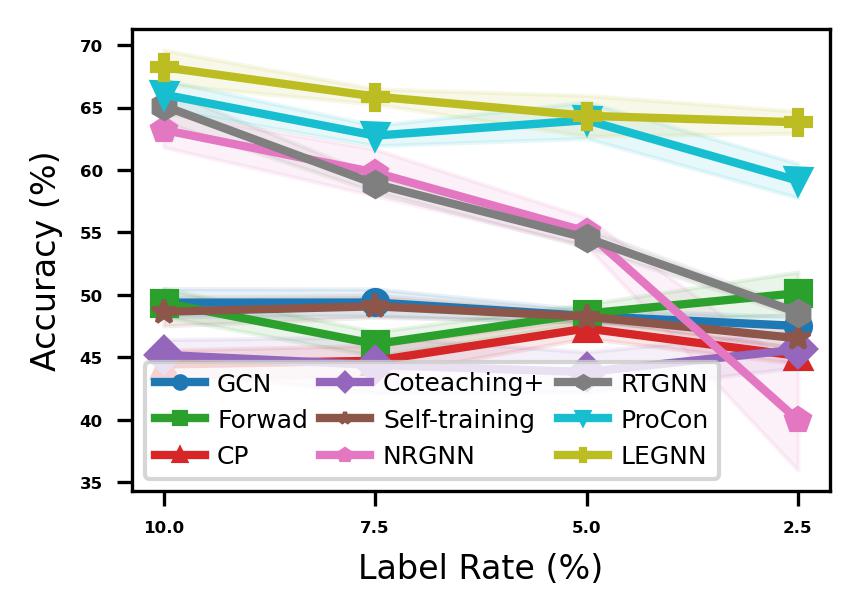}
    \end{minipage}
    \begin{minipage}{0.4\linewidth}
        \centering
        \includegraphics[width=0.99\columnwidth]{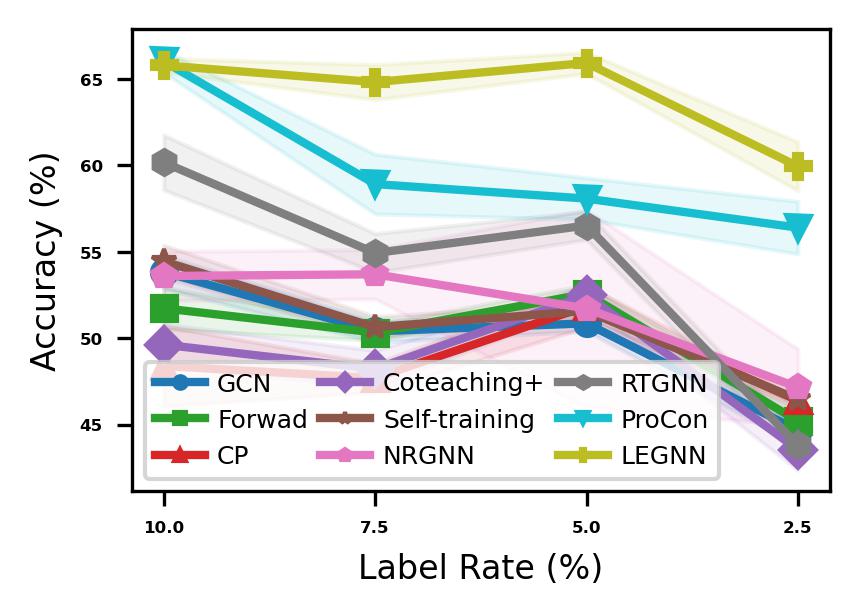} 
    \end{minipage}
    \caption{The average classification accuracy decreases as the number of labels decreases. Experiments were conducted on the Citeseer dataset, with the left figure representing the Sym-50\% noise and the right figure representing the Pair-40\% noise. The error bars of standard deviation are shaded in each plot.}
    \label{fig:label rate}
\end{figure}

\begin{table}[ht]
    \centering
    \resizebox{\linewidth}{!}{
    \begin{tabular}{cccccc}
        \toprule
         & $k$ & $2k$ & $3k$ & $4k$ & $5k$ \\
        \midrule
        Citeseer & 6.54\% & 12.74\% & 18.43\% & 23.88\% & 29.16\%\\
        Pubmed & 1.57\% & 3.06\% & 4.51\% & 5.96\% & 7.36\%\\
        \bottomrule
    \end{tabular}
    }
     \caption{Comparison of edge construction performance for NRGNN.}
    \label{tab:edgecons}
\end{table}

Furthermore, we compared the mean performance of LEGNN and baseline methods under different label rates when higher noise rates were applied (Sym-50\% and Pair-40\%), as illustrated in Fig. \ref{fig:label rate}. Consistent with RTGNN, when the label rate is $r$, we set the proportion of the validation set to be $20\% - r$. The shaded regions in the figure represent the standard deviations. It is evident from the graph that as the label rate decreases, LEGNN's degradation rate is much slower compared to the contrast methods. This phenomenon can be attributed to the label ensemble strategy, which provides abundant labels, effectively aiding the network in acquiring rich supervision with a limited number of labeled samples.

\smallskip\noindent\textbf{Experiments on Large-scale Datasets (\textbf{RQ2})} To validate the scalability of LEGNN for large-scale datasets, we present the experimental results under different noise types and rates on the IGB and OGBN-Arxiv dataset in Table \ref{tab:acc_large}. In Table \ref{tab:acc_large}, the proposed method achieves the highest mean accuracy on all cases.

Taken together, the results on Tables~\ref{tab:acc_small}, ~\ref{tab:acc_large} and ~\ref{tab:Clothing1m} that LEGNN provides competitive robustness across diverse noise settings. We attribute these gains to the additional information gain brought by bootstrapping the neighbor context, which enriches the candidate label set with more label information correlated with clean labels. From the perspective of reliable labeling, imposing a more rigorous constraint on reliability can improve the quality of pseudo-labels (higher precision). However, this also leads to a significant loss of other potentially correct information (lower recall). In fact, the issue of sacrificing recall for precision in pseudo-labeling has been overlooked in previous work, and we are the first to identify and address this problem through partial label learning techniques, providing new insights into pseudo-label construction for graph learning. LEGNN incorporates more clean labels into the model training by preserving potentially correct labels (higher recall). Although this introduces more noisy information (lower precision), the improved F1-score demonstrates a better balance between precision and recall. We provide a detailed analysis of this in Section \ref{sec:em}.

\smallskip\noindent\textbf{Discussion} \textit{Why NRGNN and RTGNN perform worse in Pubmed and OGBN.}\\
This may be because NRGNN and RTGNN use node representations to construct additional similarity edges to assist with pseudo-label assignment. When errors occur in edge construction, it can affect the accuracy of the pseudo labels, leading to further error accumulation. Under the influence of incorrect pseudo labels and noisy edges (both neighbors and labels are affected), the performance of NRGNN and RTGNN may be worse than that of baselines that do not construct additional edges (where at least the neighbors are correct). For Pubmed and OGBN, both datasets share the characteristic of having lower feature dimensions, which makes inferring potential edges from nodes more challenging. To further demonstrate that the poor performance of edge-based construction strategies on OGBN and Pubmed is related to ineffective edge construction, we conducted an experimental analysis of NRGNN's edge construction under the Sym-50\% noise setting. Specifically, we compared the edges constructed by NRGNN with the benchmark edges. Here, we used a clean and fully supervised GCN to extract features and constructed neighboring edges based on these features as benchmarks. Table \ref{tab:edgecons} shows the proportion of edges constructed by NRGNN that match the nearest $[k, 2k, 3k, 4k, 5k]$ neighboring benchmark edges, where $k$ represents the number of edges constructed by NRGNN. In the table, we compare the edge construction performance on the Pubmed dataset (where NRGNN performs relatively poorly) and the Citeseer dataset (where NRGNN shows competitive performance). For the Pubmed dataset, the proportion of NRGNN correctly identifying nearest $k$ neighboring edges is only 1.57\%, and for the nearest $5k$ neighboring edges, it is just 7.36\%. This low hit rate negatively impacts subsequent model training. In contrast, for the Citeseer dataset, the hit rate for the nearest $5k$ neighboring edges is 29.16\%, and the proportion for the nearest k neighboring edges exceeds 6\%, resulting in good performance on Citeseer. Therefore, the reason these baselines are competitive on some datasets but less so on others is due to their differing technical focuses. Our method avoids this issue by aggregating information from graph views with diverse edges.

\subsection{Comparison of Complexity}
To address \textbf{RQ3}, we compared LEGNN with representative noise-robust GNN methods. Table \ref{tab:time} reports the average running time over five trials, total time complexity, and space complexity, with GCN as the backbone reference. LEGNN achieves the shortest running time across all five datasets, providing a $4.63$--$60.56\times$ speedup over the fastest noise-robust baseline on each dataset. These results are consistent with our analysis in Section \ref{sec:complexity}: with the remaining parameters fixed, LEGNN's total time and space complexities scale as $O(N+|E|)$, whereas the comparison methods introduce quadratic or higher node-dependent time terms and quadratic space terms. This demonstrates LEGNN's computational advantage for robust learning on large-scale sparse graphs.

\begin{table*}[htb]
    \centering
    \resizebox{0.99\textwidth}{!}{
    \begin{tabular}{cccccccc}
        \toprule
        & Cora & Citeseer & Pubmed & IGB & OGBN & Total Time Complexity & Space Complexity\\
        \midrule
        GCN (backbone) & 3.19 & 3.02 & 3.80 & 3.92 & 3.89
        & $O(TLd^2N+TLd|E|)$
        & $O(LdN+|E|+Ld^2)$ \\

        NRGNN & 28.81 & 30.10 & 181.01 & 1109.01 & 1889.33
        & $O((F+\log N+TLd)N^2+T(Ld^2+Kd+C)N+(N+TLd)|E|)$
        & $O(N^2+(Ld+K+C)N+|E|+Ld^2)$ \\

        RTGNN & 92.68 & 73.30 & 585.22 & 3521.11 & 2602.06
        & $O((F+\log N)N^2+T[Ld^2+K(Ld+C)+\log N]N+T(Ld+C)|E|)$
        & $O(N^2+(Ld+K)N+|E|+Ld^2)$ \\

        ProCon & 33.37 & 34.53 & 39.27 & 463.54 & 1780.37
        & $O(N^3+FN^2+T(Ld^2+Cd+K+I)N+TLd|E|)$
        & $O(N^2+(Ld+K+C)N+|E|+Ld^2)$ \\

        LEGNN & \textbf{5.05} & \textbf{4.81} & \textbf{8.48} & \textbf{60.27} & \textbf{29.40}
        & $O((T+RM_e)(Ld^2+C)N+(T+RM_e)Ld|E|)$
        & $O((Ld+C)N+|E|+Ld^2)$ \\
        \bottomrule
    \end{tabular}
    }
    \caption{Average runtime (seconds) over five trials, total time complexity, and space complexity.
    The shortest time excluding GCN is in \textbf{bold}.
    $N$, $|E|$, $L$, and $C$ denote node, edge, GNN layer, and class counts;
    $F$ and $d$ denote input and hidden dimensions;
    $T$ denotes training epochs, including initialization.
    $R$ and $M_e$ denote LEGNN's label generation rounds and masks per round.
    $K$ denotes per-node neighbor/negative-sampling budgets in NRGNN/RTGNN or peer count in ProCon;
    $I$ denotes ProCon's GMM iterations.
    Bounds assume uniform-width GNNs and dense two-hop computation in ProCon;
    space excludes shared input features.}
    \label{tab:time}
\end{table*}

\subsection{Analysis of Gathered Labels} \label{sec:em}
For LEGNN, a natural question is why the constructed label set, which also introduces more noise, does not lead to reduced performance (\textbf{RQ4})? In fact, this is because LEGNN better balances label precision and recall. To illustrate this phenomenon, we conducted experiments to analyze the quality of the gathered labels in Table \ref{tab:label analysis}. 

\begin{table}[htb]
\centering
\resizebox{0.99\columnwidth}{!}{
\begin{tabular}{cccccccccccccccc}
    \toprule
    \multirow{2}*{Method} & 
    \multicolumn{3}{c}{0.1} & 
    \multicolumn{3}{c}{0.2} & 
    \multicolumn{3}{c}{0.3} & 
    \multicolumn{3}{c}{0.4} & 
    \multicolumn{3}{c}{0.5} \\
    \cmidrule(lr){2-4}\cmidrule(lr){5-7}\cmidrule(lr){8-10}\cmidrule(lr){11-13}\cmidrule(lr){14-16}
     & P & R & F1 & P & R & F1 & P & R & F1 & P & R & F1 & P & R & F1 \\
    \midrule
    RTGNN & 85.71 & 18.48 & 30.41 & 85.52 & 15.12 & 25.69 & 80.85 & 12.61 & 21.81 & 75.83 & 8.63 & 15.49 & 75.73 & 9.76 & 17.30 \\
    Propagation & 42.84 & 75.45 & 54.65 & 44.82 & 77.73 & 56.86 & 41.74 & 74.88 & 53.60 & 43.98 & 75.55 & 55.60 & 38.90 & 66.26 & 49.02 \\
    LEGNN & 66.84 & 80.71 & \textbf{73.12} & 69.61 & 78.15 & \textbf{73.63} & 64.67 & 75.21 & \textbf{69.54} & 67.17 & 77.39 & \textbf{71.92} & 63.54 & 72.70 & \textbf{67.81} \\
    \bottomrule
    \end{tabular}
    }
    \caption{Label construction performance (\%) across noise rates: Precision (P), Recall (R), and F1-score.}
    \label{tab:label analysis}
\end{table}

\smallskip\noindent\textbf{Balance Between Precision and Recall of Labels} In Table \ref{tab:label analysis}, we compared the effectiveness of correction-based approach (RTGNN), propagation-based (Propagation) approach and the ensemble-based approach (LEGNN) for generating labels on the Citeseer benchmark under symmetric noise. For Propagation, we iteratively aggregate labels for each node from its neighbors for training with proposed weighted bidirectional loss, and then updated the labels based on the trained model. The other settings is consistent with LEGNN. The overall process is illustrated in the Supplementary Material Section 4.1. For Propagation and LEGNN, we treat each label in a multi-label as an independent label for calculating precision and recall. In Table \ref{tab:label analysis}, we can observe that:
\begin{itemize}[leftmargin=*, itemsep=0pt, topsep=0pt]
	\item Propagation and LEGNN have similar recall, but LEGNN exhibits much higher precision. This arises because Propagation gathers labels from all neighboring nodes, while LEGNN pays more attention to node labels within similar neighbors, mitigating the influence of dissimilar neighbors.
	\item RTGNN demonstrates better precision; however, this precision is achieved by selecting small scale training labels, resulting in significantly lower recall.
	\item From the F1-scores, it is evident that LEGNN achieves a more balanced trade-off between precision and recall. 
\end{itemize}

In summary, despite a slight decrease in precision (or an increase of noise), LEGNN allows more labeled samples to participate in training. This is because the reliable labeling methods struggle to determine the correct labels and are thus compelled to select high-confidence samples. In contrast, LEGNN constructs the label set by gathering all potentially correct labels for each node. Additionally, Table \ref{tab:labelcole} reports clean-label coverage. Specifically, we compare the proportion of true labels contained in the labels provided by the LEGNN and the basic confidence-based reliable labeling method (Confidence) on the Citeseer dataset. For Confidence, we iteratively select samples with model prediction probabilities above 0.99 for training a network with a cross-entropy loss, and update remaining labels based on the trained model. The overall process is illustrated in the Supplementary Material Section 4.2. The table shows that LEGNN contains more clean labels. Even as the noise rate increases, the number of clean labels provided by LEGNN does not significantly decrease. This indicates that the gathered labels contain more useful supervised information. 

\begin{table}[htb]
    \centering
    \resizebox{0.99\columnwidth}{!}{
    \begin{tabular}{cccccc}
        \toprule
        Method & Sym-10\% & Sym-20\% & Sym-30\% & Sym-40\% & Sym-50\% \\
        \midrule
        LEGNN & \textbf{79.99$\pm$0.65} & \textbf{79.65$\pm$0.39} &\textbf{77.85$\pm$1.74} &\textbf{74.54$\pm$4.29} & \textbf{72.70$\pm$5.41}\\
        Confidence & 75.04$\pm$0.77&73.87$\pm$1.01&68.55$\pm$3.01&67.29$\pm$3.37&57.79$\pm$2.97\\
        \bottomrule
    \end{tabular}}
     \caption{Comparison of clean label coverage.}
    \label{tab:labelcole}
\end{table}

\subsection{Ablation Studies and Component Analysis}

To address \textbf{RQ5}, we conducted ablation experiments on components of LEGNN and performed analysis on the parameters. Specifically, we first analyzed the impact of different components, and then systematically examined how various backbone networks, masking strategies, and parameter choices affect LEGNN's performance.

\smallskip\noindent\textbf{Ablation} Table \ref{tab:ablation} presents ablation results on Citeseer under Sym-50\% noise. In the table, we report results for removing the low-probability labels (w/o low-probability), removing the high-probability labels (w/o high-probability), removing the label gathering strategy (w/o gathering), and removing both low-probability labels and the label gathering strategy. Here, when low-probability labels are removed, Eq. \eqref{eq:loss} degenerates to Eq. \eqref{eq:positive}; when the label gathering strategy is removed, the method becomes label correction with unweighted bidirectional loss; and when both are removed, the method reduces to the basic label correction approach. Here, GCN serves as our backbone network. The results demonstrate that the proposed label gathering module and the additional low-probability labels effectively enhance the model's performance. It should be noted that, w/o high-probability, where only low-probability labels are used, results in a significant performance drop. This is because low-probability labels, despite having lower errors, provide less information. For a $C$-class classification task, the information from $C-1$ incorrect labels is equivalent to that from $1$ correct label. This explains why, as discussed in Section \ref{sec:ensemble}, we use low-probability labels only as a supplement to smooth out high-probability label errors.

\begin{table}[htb]
\centering
\resizebox{0.98\columnwidth}{!}{
    \begin{tabular}{lcccccc}
    \toprule
    Metric & LEGNN & w/o low-prob. & w/o high-prob. & w/o gathering & w/o gather.\&low-prob. & GCN \\
    \midrule
    Accuracy (\%) & \textbf{69.62$\pm$3.69} & 67.06$\pm$3.99 & 31.54$\pm$2.19 & 55.49$\pm$1.86 & 55.47$\pm$1.91 & 46.78$\pm$0.70 \\
    \bottomrule
    \end{tabular}
    }
\caption{Ablation experiments on model components.}
\label{tab:ablation}
\end{table}

\smallskip\noindent\textbf{Training with Different Masking Strategies} We design an efficient GNN training process by employing a lower complexity but less robust random masking strategy. As shown in Table \ref{tab:masking}, we conducted additional experiments on Citeseer with three variants: (a)LEGNN-N: node similarity-based masking by removing the nearest neighbors; (b)LEGNN-W: weight-based method, where edges are sampled for masking according to a probability guided by their weights, calculated from similarity. (c) LEGNN-S: structural masking with degree-adaptive probabilities, where edges are sampled for masking based on the normalized degree values of their connected nodes. These reasonable strategies demonstrate improved prediction accuracy under most noisy label conditions compared to baseline methods, especially in pair flipping. Nevertheless, our technical emphasis remains on balancing computational efficiency with noise robustness. As quantified in Table \ref{tab:masking}, the advanced masking strategies (LEGNN-N/W/S) incur 3-4× longer training times compared to random masking — a consequence of their $O(N^2)$ level complexity from pairwise feature similarity computations (LEGNN-N/W) and pairwise degree summations (LEGNN-S). This contrasts sharply with our baseline random masking's $O(|E|)$ level complexity, which scales linearly with edges.

\begin{table}[htb]
    \centering
    \resizebox{0.95\linewidth}{!}{
        \begin{tabular}{cccccc}
            \toprule
            ~ & Sym-20\% & Sym-50\% & Pair-20\% & Pair-40\% & Running Time\\ 
            \midrule
            LEGNN & 74.70±0.87 & \textbf{69.62±3.69} & 73.38±1.64 & 64.82±2.20 & \textbf{4.81} \\
            LEGNN-N & \textbf{75.96±0.28} & 69.58±3.13 & 74.56±0.96 & 67.43±1.99 & 23.12 \\
            LEGNN-W & 75.18±0.63 & 66.84±2.66 & 75.65±0.47 & 70.93±2.69 & 24.44 \\
            LEGNN-S & 75.49±0.95 & 66.65±3.67 & \textbf{75.65±0.38} & \textbf{72.23±2.30} & 19.79 \\
            \bottomrule
        \end{tabular}
    }
    \caption{The average classification accuracy and standard deviation (in percentage) with different maskings.}\label{tab:masking}
\end{table}

\smallskip\noindent\textbf{Training with Different Backbones} To ensure the efficiency of our method, we used a basic two-layer GCN as the backbone in comparison experiments. To further demonstrate the effectiveness of our method for training different GNN model against label noise, we supplemented the experiments with GCN (referred to as LEGNN-GCN), GAT (referred to as LEGNN-GAT), and GraphSAGE (referred to as LEGNN-GraphSAGE) backbones on the Cora, Citeseer, and OGBN-Arxiv datasets, as shown in Table \ref{tab:backbone}. On smaller datasets (i.e., Cora, Citeseer), there is no significant difference among different backbones for LEGNN. However, on larger-scale datasets (i.e., OGBN-Arxiv), LEGNN-GAT demonstrated advantages. One possible reason for this phenomenon is that on smaller datasets such as Cora and Citeseer, the graph structures and the information they carry are relatively simpler, making the differences between backbones less pronounced. In contrast, for larger datasets, GAT's ability to dynamically weight the importance of neighboring nodes becomes more valuable. Additionally, the table shows that LEGNN combined with different backbones consistently achieves better classification performance, effectively enhancing their robustness to label noise.

\begin{table*}[htb]
    \centering
    \setlength{\tabcolsep}{14pt}
    \begin{tabular}{cccccc}
        \toprule
        ~ & ~ & Sym-20\% & Sym-50\% & Pair-20\% & Pair-40\% \\ 
        \midrule
        \multirow{6}*{Cora} & GCN & 71.47±1.66 & 53.05±3.89 & 76.00±1.11 & 58.08±1.51 \\
        & LEGNN-GCN & \underline{\textbf{79.95±1.56}} & \underline{67.93±3.81} & \underline{79.64±1.00} & \underline{67.55±0.64} \\
        & GAT & 72.12±3.08 & 56.37±5.88 & 69.05±2.18 & 53.75±1.78 \\
        & LEGNN-GAT & \underline{79.08±0.86} & \underline{\textbf{68.86±2.75}} & \underline{\textbf{79.86±1.86}} & \underline{\textbf{69.43±3.45}} \\ 
        & GraphSAGE & 70.99±1.80 & 54.83±5.13 & 71.11±3.70 & 55.42±2.80 \\
        & LEGNN-GraphSAGE & \underline{78.30±0.90} & \underline{67.45±4.52} & \underline{78.99±0.96} & \underline{66.07±4.01} \\ 
        \midrule
        \multirow{6}*{Citeseer} & GCN & 62.73±1.40 & 46.78±0.70 & 66.37±1.02 & 49.91±0.96 \\
        & LEGNN-GCN & \underline{\textbf{74.70±0.87}} & \underline{69.62±3.69} & \underline{73.38±1.64} & \underline{64.82±2.20} \\
        & GAT & 65.47±4.46 & 49.87±5.78 & 66.47±3.36 & 50.17±4.08 \\ 
        & LEGNN-GAT & \underline{74.08±0.96} & \underline{\textbf{70.08±3.34}} & \underline{73.73±1.27} & \underline{\textbf{65.41±4.96}} \\ 
        & GraphSAGE & 63.20±2.73 & 50.78±5.16 & 65.72±3.72 & 51.08±5.64 \\
        & LEGNN-GraphSAGE & \underline{74.14±3.24} & \underline{69.31±5.88} & \underline{\textbf{73.90±1.48}} & \underline{65.30±2.48} \\ 
        \midrule
        \multirow{6}*{OGBN} & GCN & 46.61±1.40 & 45.43±0.73 & 46.37±1.05 & 44.94±0.81 \\
        & LEGNN-GCN & \underline{61.08±0.22} & \underline{57.48±0.43} & \underline{59.45±3.72} & \underline{58.39±0.22} \\ 
        & GAT & 62.38±0.53 & 60.75±0.66 & 61.72±0.87 & 57.85±0.22 \\
        & LEGNN-GAT & \underline{\textbf{65.57±0.36}} & \underline{\textbf{65.35±0.23}} & \underline{\textbf{64.97±0.40}} & \underline{\textbf{62.96±0.24}} \\
        & GraphSAGE & 61.91±0.21 & 60.33±1.58 & 60.06±2.20 & 57.17±0.66 \\
        & LEGNN-GraphSAGE & \underline{64.73±0.33} & \underline{64.14±0.21} & \underline{64.68±0.19} & \underline{62.74±0.35} \\ 
        \bottomrule
    \end{tabular}
    \caption{The average classification accuracy and standard deviation (expressed in percentage) across five trials of different backbones. Results superior to the baseline are marked with underscores, while the best results are highlighted in \textbf{bold}.} \label{tab:backbone}
\end{table*}

\begin{figure}[htb]
    \centering
    \begin{minipage}{0.4\linewidth}
        \centering
        \subfloat[Sym-20\%\label{fig:0.2param}]{\includegraphics[width=\linewidth]{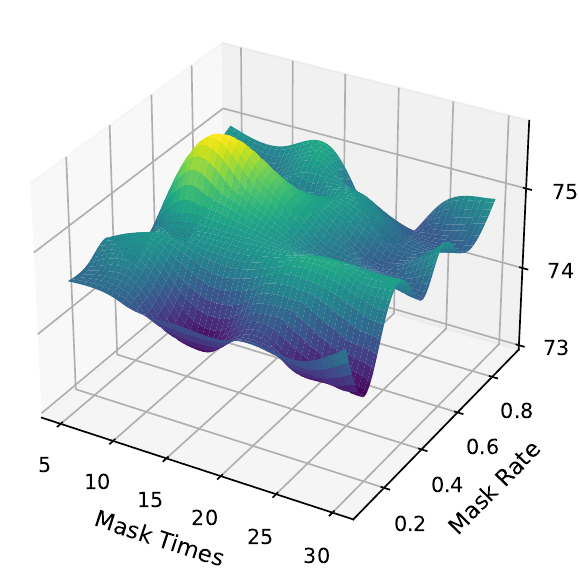}}
    \end{minipage}
    \begin{minipage}{0.4\linewidth}
        \centering
        \subfloat[Sym-50\%\label{fig:0.5param}]{\includegraphics[width=\linewidth]{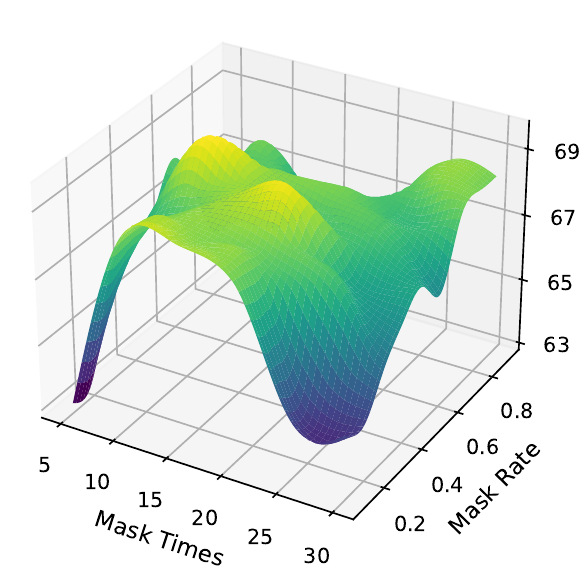}}
    \end{minipage}
    \caption{The average accuracy (\%) across five trials with different parameters on Citeseer.}
    \label{fig:params}
\end{figure}

\smallskip\noindent\textbf{Parameter Analysis} Two primary hyperparameters influence the effectiveness of this study: the neighbor mask rate and the number of masked graphs. We varied the mask rate $\{0.1, \dots, 0.9\}$ and mask times $\{5, 10, \dots, 30\}$, respectively, and reported the accuracy in Fig. \ref{fig:params}. We observe an inverse relationship between mask rate and mask times: as the mask rate increases, the optimal mask times decrease, and vice versa, particularly at a 50\% noise rate. This occurs because a higher mask rate increases label diversity in each masking step, requiring fewer mask times to maintain stability, while a lower mask rate requires more mask times to achieve sufficient diversity.

To further understand the sensitivity of these parameters across different conditions, we conducted an analysis of parameter sensitivity with respect to different noise rates, graph densities (controlled by a density level parameter that determines the proportion of retained edges), and class imbalances (with an imbalance level parameter $\rho$, where class $i$ has a retention ratio of $\rho^i$), as shown in Table \ref{tab:param_combined}. Across different noise rates, the relationship between mask rate and mask times remains consistent with previous analysis, with no significant changes across noise levels. For graph densities, as sparsity increases, more conservative choices of mask rate and mask times yield better performance, as fewer edges require a more cautious strategy to retain information and ensure stable label predictions. For class imbalance, as imbalance increases, both the mask rate and mask times also increase. This is because our label ensemble strategy can be viewed as an upsampling of labels, and higher imbalance levels require stronger upsampling to mitigate the imbalance. 

\begin{table*}[htb]
\centering
\resizebox{\textwidth}{!}{
\begin{tabular}{cccccccccccccccc}
    \toprule
    &\multicolumn{5}{c}{\textbf{Noise Rate}} & \multicolumn{5}{c}{\textbf{Graph Density}} & \multicolumn{5}{c}{\textbf{Class Imbalance}} \\
    \cmidrule(lr){2-6}\cmidrule(lr){7-11}\cmidrule(lr){12-16}
     & 10\% & 20\% & 30\% & 40\% & 50\% & 0.9 & 0.7 & 0.5 & 0.3 & 0.1 & 0.1 & 0.2 & 0.3 & 0.4 & 0.5 \\
    \midrule
    Accuracy (\%) & 76.72 & 76.06 & 74.98 & 75.24 & 70.02 & 75.55 & 73.99 & 72.09 & 67.56 & 64.23 & 75.84 & 74.45 & 73.91 & 67.16 & 59.47 \\
    Optimal Mask Rate & 0.2 & 0.9 & 0.3 & 0.1 & 0.2 & 0.5 & 0.6 & 0.6 & 0.2 & 0.3 & 0.1 & 0.5 & 0.4 & 0.7 & 0.7 \\
    Optimal Mask Times & 25 & 10 & 15 & 25 & 20 & 20 & 5 & 5 & 5 & 5 & 30 & 20 & 15 & 25 & 20 \\
    \bottomrule
\end{tabular}}
\caption{Optimal setting of mask times and mask rate on Citeseer under different conditions.}
\label{tab:param_combined}
\end{table*}

To substantiate this claim, we provide an empirical evidence on Cora, Citeseer, and Pubmed in Table \ref{tab:imba_count} (shown in the next page), as these datasets' label set are relatively small, making imbalance more apparent. In the table, we present the proportion and variance of noisy label nodes in each class. The results show that, under most benchmarks, our strategy effectively reduces class imbalance. A specific example is the Cora Pair-20\% case, where original noise label distribution is [0.0909, 0.1818, 0.0455, 0.0909, 0.2273, 0.2727, 0.0909]. After applying our strategy, the distribution changes to [0.1364, 0.1364, 0.1818, 0.0909, 0.1364, 0.1591, 0.1591], with the smaller classes 1 and 4 noticeably increased. The initial class distribution showed high variance ($\sigma^2=0.0072$), which was significantly reduced after applying our ensemble strategy ($\sigma^2=0.0008$), demonstrating an 89\% reduction.

\begin{table}[htb]
    \centering
    \resizebox{0.99\linewidth}{!}{
        \begin{tabular}{cccccccc}
            \toprule
            ~ & ~ & Sym-20\% & Sym-50\% & Pair-20\% & Pair-40\% & IDN-20\% & IDN-50\%\\ 
            \midrule
            \multirow{2}*{Cora} & Before & 0.0072 & 0.0390 & 0.0117 & 0.0140 & \textbf{0.0053} & 0.0027 \\
            & Ensemble & \textbf{0.0008} & \textbf{0.0020} & \textbf{0.0068} & \textbf{0.0087} & 0.0069 & \textbf{0.0024} \\
            \midrule
            \multirow{2}*{Citeseer} & Before & 0.0042 & 0.0079 & 0.0191 & 0.0126 & 0.0039& 0.0053 \\
            & Ensemble & \textbf{0.0034} & \textbf{0.0045} & \textbf{0.0151} & \textbf{0.0110} & \textbf{0.0019} & \textbf{0.0048} \\
            \midrule
            \multirow{2}*{Pubmed} & Before & 0.0100 & \textbf{0.0014} & 0.0003 & 0.0041 & 0.0208 & \textbf{0.0030} \\
            & Ensemble & \textbf{0.0092} & 0.0063 & \textbf{0.0001} & \textbf{0.0011} & \textbf{0.0081} & 0.0072 \\
            \bottomrule
        \end{tabular}
    }
    \caption{The variances before and after label ensemble operation.}\label{tab:imba_count}
\end{table}

\section{Conclusion}
We have presented LEGNN, a reduced-complexity framework for graph node classification under label noise from a label-ensemble perspective. Instead of aggressively filtering to a single ``clean'' pseudo-label, LEGNN aggregates diverse predictions from masked, bootstrapped neighborhood contexts into a candidate label set and optimizes a partial-label objective over this set. This construction preserves plausible ground-truth labels that strict filtering tends to discard, improving the quality of the constructed label sets (e.g., higher F1 against ground truth in our analysis) and yielding a better precision–recall balance of pseudo-label supervision. To further enhance robustness when the candidate set still contains noise, LEGNN employs a symmetric objective that leverages low-probability predictions as negative signals, which helps counteract error accumulation. Experiments show that LEGNN achieves competitive performance while demonstrating favorable empirical scaling in training time, avoiding quadratic-cost computations and heavy auxiliary modules. Consequently, the framework can be incorporated with minor changes into existing GNN training pipelines to improve robustness on large-scale noisy graphs.

While LEGNN achieves a favorable trade-off between efficiency and robustness, two limitations merit further study. First, the bootstrapping relies on random neighbor masking. This strategy encourages diverse neighborhood views, yet it remains a structure-agnostic heuristic that treats neighbors uniformly. Second, the label-set construction draws its evidence strictly from local neighborhoods; for very low-degree nodes, the available context is limited, which can make the constructed label sets less informative than those for well-connected nodes. Future work will focus on refining the bootstrapping mechanism by developing low-complexity, structure- or confidence-aware masking rules (e.g., via lightweight attention or uncertainty-guided selection) and exploring inexpensive context expansion (e.g., conservative topology augmentation or controlled multi-hop aggregation) to support low-degree nodes. Beyond these algorithmic improvements, extending the label ensemble perspective to more complex graph scenarios (e.g., dynamic graphs) is another promising direction.

\section*{Acknowledgments}
This research was partially supported by the Key Research and Development Project in Shaanxi Province No. 2024PTZCK-89, and the National Natural Science Foundation of China No. 62406242, 62672397 and 62476215.

\bibliographystyle{elsarticle-num} 
\bibliography{ref}

\end{document}